\documentclass[runningheads]{llncs}
\usepackage[latin9]{inputenc}
\usepackage{textcomp}
\usepackage{amsmath}
\usepackage{amssymb}
\usepackage{graphicx}
\usepackage{wasysym}

\makeatletter

\providecommand{\tabularnewline}{\\}
\newcommand{\lyxdot}{.}

\usepackage{comment}
\usepackage{color}
\usepackage{rotating}

\usepackage{tikz}

\newcommand\blfootnote[1]{%
	\begingroup
	\renewcommand\thefootnote{}\footnote{#1}%
	\addtocounter{footnote}{-1}%
	\endgroup
}

\mainmatter 
\def\ECCVSubNumber{2437}  
\titlerunning{CurveLane-NAS}  
\authorrunning{Hang Xu et al.}  
\author{Hang Xu\inst{*}\inst{1}, Shaoju Wang\inst{*}\inst{2}, Xinyue Cai\inst{1},\\ Wei Zhang\inst{1}, Xiaodan Liang\inst{\dagger}\inst{2}, Zhenguo Li\inst{1}} 
\institute{Huawei Noah's Ark Lab \and
	Sun Yat-sen University}
\makeatother

\begin{document}
\title{CurveLane-NAS: Unifying Lane-Sensitive Architecture Search and Adaptive
 Point Blending}
\maketitle
\begin{abstract}
\blfootnote{* Equally Contributed.}
\blfootnote{$\dagger$ Corresponding Author: xdliang328@gmail.com}
We address the curve lane detection problem which poses more realistic
challenges than conventional lane detection for better facilitating
modern assisted/autonomous driving systems. Current hand-designed
lane detection methods are not robust enough to capture the curve
lanes especially the remote parts due to the lack of modeling both
long-range contextual information and detailed curve trajectory. In
this paper, we propose a novel lane-sensitive architecture search
framework named CurveLane-NAS to automatically capture both long-ranged
coherent and accurate short-range curve information. It consists of three search modules: a) a feature
fusion search module to find a better fusion of the local and global
context for multi-level hierarchy features; b) an elastic backbone
search module to explore an efficient feature extractor with good
semantics and latency; c) an adaptive point blending module to search
a multi-level post-processing refinement strategy to combine multi-scale
head prediction. Furthermore,
we also steer forward to release a more challenging benchmark named
CurveLanes for addressing the most difficult curve lanes. It consists
of 150K images with 680K labels.\footnote{The new dataset can be downloaded at http://www.noahlab.com.hk/opensource/\\vega/\#curvelanes.} Experiments on the new CurveLanes show that the SOTA lane detection
methods suffer substantial performance drop while our model can still
reach an 80+\% F1-score. Extensive experiments on traditional lane
benchmarks such as CULane also demonstrate the superiority of our
CurveLane-NAS, e.g. achieving a new SOTA 74.8\% F1-score on CULane.

\end{abstract}

\section{Introduction}

\begin{figure}

\begin{centering}
\includegraphics[width=1\columnwidth]{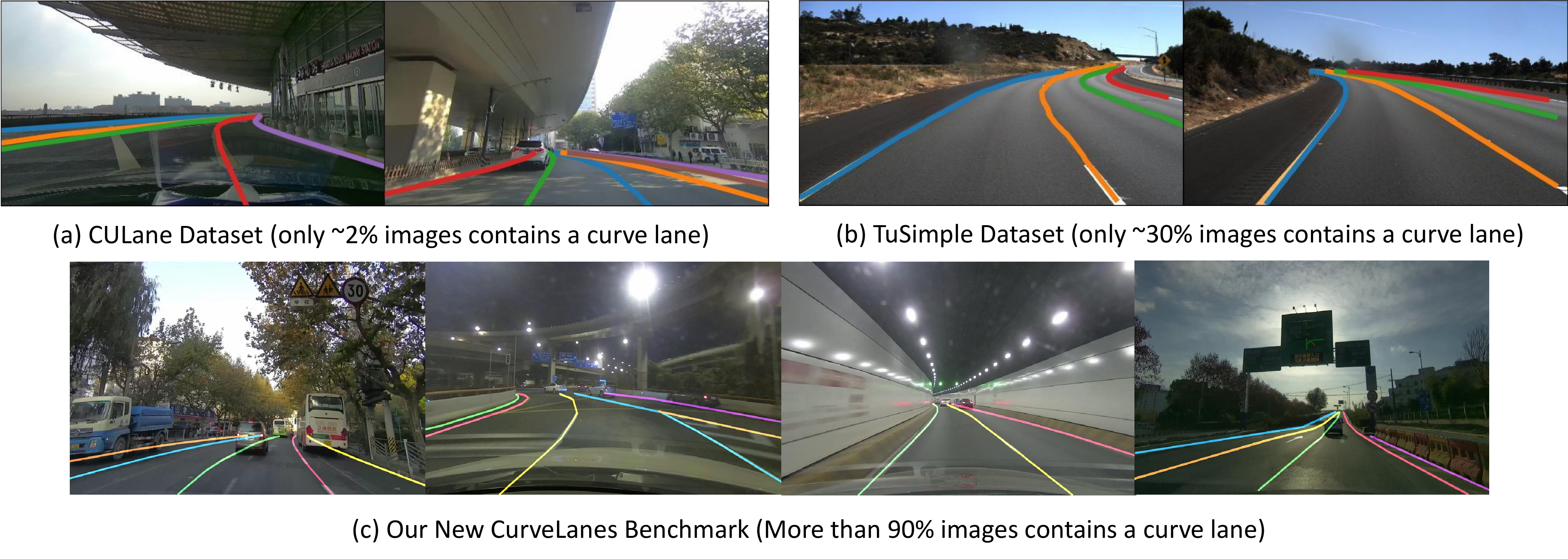}
\par\end{centering}

\caption{\label{fig:Curve-lane-detection}Examples of curve lane detection.
Comparing to straight lane, the detection of \textbf{curve lanes}
is more crucial for trajectory planning in modern assisted and autonomous
driving systems. However, the proportion of curve lanes images in
current large-scale datasets is very limited, 2\% in CULane Dataset
(around 2.6K images) and 30\% in TuSimple Dataset (around 3.9K images),
which hinders the real-world applicability of the autonomous driving
systems. Therefore, we establish a more challenging benchmark named
CurveLanes for the community. It is the largest lane detection dataset
so far (150K images) and over 90\% images (around 135K images) contain curve lane.}

\end{figure}

Lane detection is a core task in modern assisted and autonomous driving
systems to localize the accurate shape of each lane in a traffic scene.
Comparing to straight lane, the detection of curve lanes is more crucial
for further down-streaming trajectory planning tasks to keep the car
properly position itself within the road lanes during steering in
complex road scenarios. As shown in Figure \ref{fig:Curve-lane-detection},
in real applications, curve lane detection could be very challenging
considering the long varied shape of the curve lanes and likely occlusion
by other traffic objects. Furthermore, the curvature of the curve
lane is greatly increased for remote parts because of interpolation
which makes those remote parts hard to be traced. Moreover, real-time
hardware constraints and various harsh scenarios such as poor weather/light
conditions \cite{pan2018spatial} also limit the capacity of models.

Existing lane detection datasets such as TuSimple \cite{TuSimple}
and CULane \cite{pan2018spatial} are not effective enough to measure
the performance of curve lane detection. Because of the natural distribution
of lanes in traffic scenes, most of the lanes in those datasets are
straight lanes. Only about 2.1\% of images in CULane (around 2.6K),
and 30\% in TuSimple contain curve lanes (around 3.9K). To better
measure the challenging curve lane detection performance and facilitate
the studies on the difficult road scenarios, we introduce a new large-scale
lane detection dataset named CurveLanes consisting of 150K images
with carefully annotated 680K curve lanes labels. All images are carefully
picked so that almost all of the images contain at least one curve
lane (more than 135K images). To our best knowledge, it is the largest
lane detection dataset so far and establishes a more challenging benchmark
for the community.

The most state-of-the-art lane detection methods are CNN-based methods.
Dense prediction methods such as SCNN \cite{pan2018spatial} and SAD
\cite{hou2019learning} treat lane detection as a semantic segmentation
task with a heavy encoder-decoder structure. However, those methods
usually use a small input image which makes it hard to predict remote
parts of curve lanes. Moreover, those methods are often limited to
detect a pre-defined number of lanes. On the other hand, PointLaneNet
\cite{chen2019pointlanenet} and Line-CNN \cite{li2019line} follow
a proposal-based diagram which generates multiple point anchor or
line proposals in the images thus getting rid of the inefficient decoder
and pre-defined number of lanes. However, the line proposals are not
flexible enough to capture variational curvature along the curve lane.
Besides, PointLaneNet \cite{chen2019pointlanenet} prediction is based
on one fixed single feature map and fails to capture both long-range
and short-range contextual information at each proposal. They suffer
from a great performance drop in predicting difficult scenarios such
as the curve or remote lanes.

In this paper, we present a novel lane-sensitive architecture search
framework named CurveLane-NAS to solve the above limitations of current
models for curve lane detection. Inspired by recent advances in network
architecture search (NAS) \cite{zoph2018learning,liu2018progressive,liu2018darts,real2019regularized,tan2019efficientnet,chen2018searching,liu2019auto},
we attempt to automatically explore and optimize current architectures
to an efficient task-specific curve lane detector. A search space
with a combination of multi-level prediction heads and a multi-level
feature fusion is proposed to incorporate both long-ranged coherent
lane information and accurate short-range curve information. Besides,
since post-processing is crucial for the final result, we unify the
architecture search with optimizing the post-processing step by adaptive
point blending. The mutual guidance of the unified framework ensures
a holistic optimization of the lane-sensitive model. Specifically,
we design three search modules for the proposal-based lane detection:
1) an elastic backbone search module to allocate different computation
across multi-size feature maps to explore an efficient feature extractor
for a better trade-off with good semantics and latency; 2) a feature
fusion search module is used to find a better fusion of the local
and global context for multi-level hierarchy features; 3) an adaptive
point blending module to search a novel multi-level post-processing
refinement strategy to combine multi-level head prediction and allow
more robust prediction over the shape variances and remote lanes.
We consider a simple yet effective multi-objective search algorithm
with the evolutionary algorithm to properly allocate computation with reasonable receptive fields and spatial
resolution for each feature level thus reaching an optimal trade-off
between efficiency and accuracy. 

\begin{figure}[t]

\begin{centering}
\includegraphics[width=1\columnwidth]{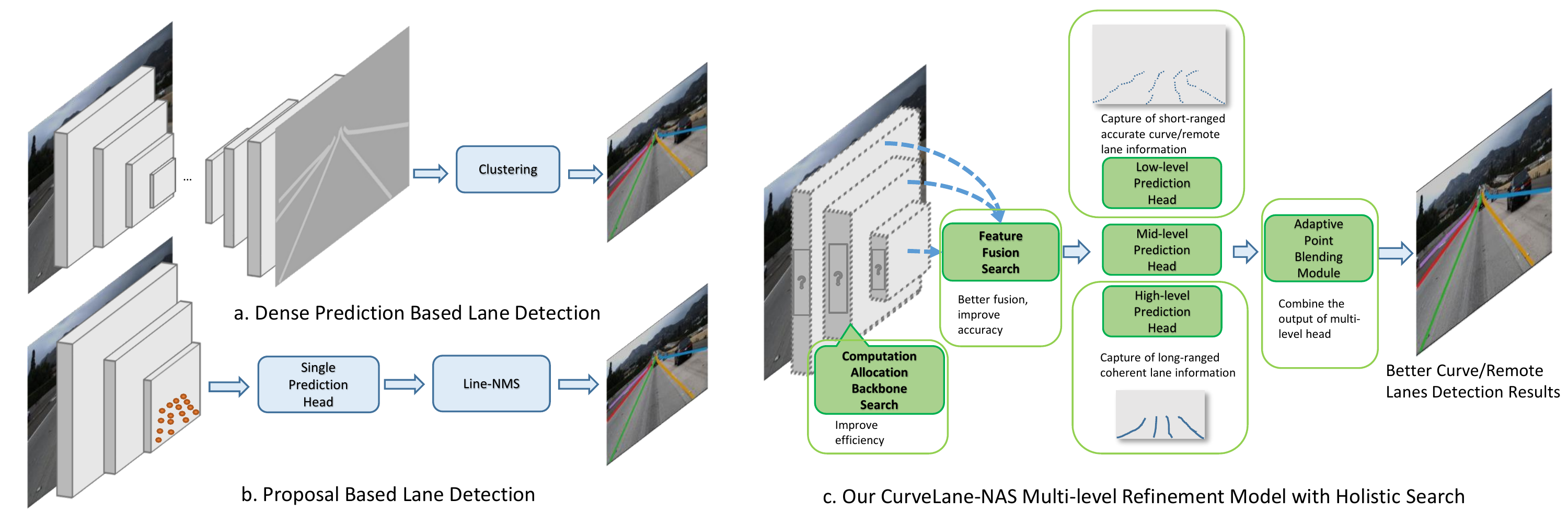}
\par\end{centering}

\caption{\label{fig:benchmark-graph}Comparison of the Lane Detection frameworks:
(a) dense prediction based (SCNN \cite{pan2018spatial}), (b) proposal
based (Line-CNN \cite{li2019line}), and (c) our CurveLane-NAS. CurveLane-NAS
is a unified neural architecture search framework to discover novel
holistic network architectures for more robust lane predictions.}

\end{figure}

Experiments on the new CurveLanes show that the state-of-the-art lane
detection methods suffer substantial performance drop (10\%\textasciitilde 20\%
in terms of F1 score) while our model remains resilient. Extensive
experiments are also conducted on multiple existing lane detection
benchmarks including TuSimple and CULane. The results demonstrate
the effectiveness of our method, e.g. the searched model outperforms SCNN \cite{pan2018spatial} and SAD \cite{hou2019learning}
and achieves a new SOTA result 74.8\% F1-score on the CULane dataset with
a reduced FLOPS.

\section{Related Work}

\textbf{Lane detection. }Lane detection aims to detect the accurate
location and shape of each lane on the road. It is the core problem
in modern assisted and autonomous driving systems. Conventional methods
usually are based on hand-crafted low-level features \cite{chiu2005lane,lee2009effective,gonzalez2000lane}.
Deep learning has then been employed to extract features in an end-to-end
manner. Most lane detection works follow pixel-level segmentation-based
approach \cite{DBLP:journals/corr/abs-1907-01294,DBLP:journals/corr/abs-1905-03704,DBLP:journals/corr/abs-1909-00798,zou2019robust}
as shown in Figure \ref{fig:benchmark-graph} (a). These approaches
usually adopt the dense prediction formulation, i.e., treat lane detection
as a semantic segmentation task, where each pixel in an image is assigned
with a label to indicate whether it belongs to a lane or not. \cite{zou2019robust}
combines a semantic segmentation CNN and a recurrent neural network
to enable a consistent lane detection. SCNN \cite{pan2018spatial}
generalizes traditional deep convolutions to slice-by-slice convolution,
thus enabling message passing between pixels across rows and columns.
However, pixel-wise prediction usually requires more computation and
is limited to detect a pre-defined, fixed number of lanes. On the
other hand, several works\textbf{ }use proposal-based approaches for
efficient lane detection as shown in Figure \ref{fig:benchmark-graph}
(b). These approaches generate multiple anchors or lines proposals
in the images. PointLaneNet \cite{chen2019pointlanenet} finds the
lane location by predicting the offsets of each anchor point. Line-CNN
\cite{li2019line} introduces a line proposal network to propose a
set of ray aiming to capture the actual lane. However, those methods
predict on one single feature map and overlook the crucial semantic
information for curve lanes and remote lanes in the feature hierarchy.

\textbf{Neural Architecture Search. }NAS aims at freeing expert's
labor of designing a network by automatically finding an efficient
neural network architecture for a certain task and dataset. Most works
search a basic CNN architectures for a classification model \cite{liu2018darts,cai2018proxylessnas,liu2018progressive,tan2018mnasnet,xie2018snas}
while a few of them focus on more complicated high-level vision tasks
such as semantic segmentation and object detection \cite{chen2018searching,chen2019detnas,xu2019auto,yao2020sm}.
Searching strategies in NAS area can be usually divided into three
categories: 1) Reinforcement learning based methods \cite{baker2016designing,zoph2018learning,cai2018efficient,zhong2018practical}
train a RNN policy controller to generate a sequence of actions to
specify CNN architecture; Zoph et al. \cite{zoph2016neural,zoph2018learning}
apply reinforcement learning to search CNN, while the search cost
is more than hundreds of GPU days. 2) Evolutionary Algorithms based
methods and Network Morphism \cite{real2017large,liu2017hierarchical,jiang2020sp}
try to \textquotedblleft evolves\textquotedblright{} architectures
by mutating the current best architectures; Real et al. \cite{real2019regularized}
introduces an age property of the tournament selection evolutionary
algorithm to favor the younger CNN candidates during the search. 3)
Gradient-based methods \cite{liu2018darts,xie2018snas,cai2018proxylessnas}
try to introduce an architecture parameter for continuous relaxation
of the discrete search space, thus allowing weight-sharing and differentiable
optimization of the architecture. SNAS \cite{xie2018snas} propose
a stochastic differentiable sampling approach to improve \cite{liu2018darts}.
Gradient-based methods are usually fast but not so reliable since
weight-sharing makes a big gap between the searching and final training.
RL methods usually require massive samples to converge thus a proxy
task is usually required. In this paper, by considering the task-specific
problems such as real-time requirement, severe road mark degradation,
vehicle occlusion, we carefully design a search space and a sample-based
multi-objective search algorithm to find an efficient but accurate
architecture for the curve lane detection problem.

\section{CurveLane-NAS framework}

In this paper, we present a novel lane-sensitive architecture search
framework named CurveLane-NAS to solve the limitations of current
models of curve lane detection. Figure \ref{fig:benchmark-graph}
(a) and (b) show existing lane detection frameworks. We extend the
diagram of (b) to our multi-level refinement model with a unified
architecture search framework as shown in Figure \ref{fig:benchmark-graph}
(c). We propose a flexible model search space with multi-level prediction
heads and multi-level feature fusion to incorporate both long-ranged
coherent lane information and accurate short-range curve information.
Furthermore, our search framework unifies NAS and optimizing the post-processing
step in an end-to-end fashion. 

\begin{figure}[t]
\begin{centering}
\includegraphics[width=1\columnwidth]{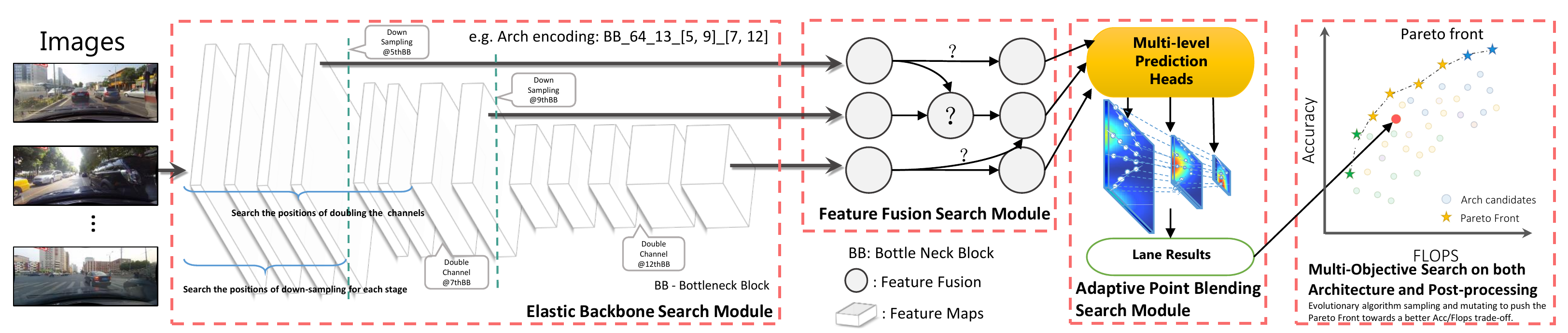}
\par\end{centering}

\caption{\label{fig:framework-graph}An overview of our NAS for lane detection
pipeline. Our unified search frameworks has three modules: 1) an elastic
backbone search module to explore an efficient feature extractor with
an optimal setting of network width, depth and when to raise channels/down-sampling,
2) a feature fusion search module to find a suitable fusion of several
feature levels; 3) an adaptive point blending module to automatically
highlight the most important regions by an adaptive masking and allow
a more robust refinement over the shape variances and remote lanes. A unified multi-objective search algorithm
is applied to generate a Pareto front with the optimal accuracy/FLOPS
trade-off.}

\end{figure}

The overview of our CurveLane-NAS framework can be found in Figure
\ref{fig:framework-graph}. We design three search modules: 1) an
elastic backbone search module to set up an efficient allocation of
computation across stages, 2) a feature fusion search module to explore
a better combination of local and global context; 3) an adaptive point
blending module to search a novel multi-level post-processing refinement
strategy and allow more robust prediction over the shape variances
and remote lanes. We consider a simple yet effective multi-objective
search algorithm to push the Pareto front towards an optimal trade-off
between efficiency and accuracy while the post-processing search can
be naturally fit in our NAS formulation. 

\subsection{Elastic Backbone Search Module}

The common backbone of a lane detection is ImageNet pretrained ResNet
\cite{he2016deep}, MoblieNet \cite{sandler2018mobilenetv2} and GoogLeNet
\cite{szegedy2015going}, which is neither task-specific nor data-specific.
The backbone is the most important part to extract relevant feature
and handcrafting backbone may not be optimal for curve lane detection.
Thus, we can resort to task-specific architecture search here to explore
novel feature extractor for a better trade-off with good semantics
and latency. 

A common backbone aims to generate intermediate-level features with
increasing down-sampling rates, which can be regarded as 4 stages.
The blocks in the same stage share the same spatial resolution. Note
that the early-stage usually has higher computational cost with more
low-level features. The late-stage feature maps are smaller thus the
computational cost is relatively smaller but losing a lot of spatial
details. How to leverage the computation cost over different stages
for an optimal lane network design? Inside the backbone, we design
a flexible search space to find the optimal base channel size, when
to down-sample and when to raise the channels as follows:

We build up the backbone by several stacked ResNet blocks \cite{he2016deep}:
basic residual block and bottleneck residual block. The backbone has
the choice of 3 or 4 stages. We allow different base channel size
${48,64,80,96,128}$ and different number of blocks in each stage
corresponding to different computational budget. The number of total
blocks variates from 10 to 45. To further allow a flexible allocation
of computation, we also search for where to raise the channels. Note
that in the original ResNet18/50, the position of doubling the channel
size block is fixed at the beginning of each stage. For example, as
shown in Figure \ref{fig:framework-graph}, the backbone architecture
encoding string looks like ``BB\_64\_13\_{[}5, 9{]}\_{[}7, 12{]}''
where the first placeholder encodes the block setting, 64 is the base
channel size, 13 is the total number of blocks and {[}5, 9{]} are
the position of down-sampling blocks and {[}7, 12{]} are the position
of doubling channel size. 

The total search space of the backbone search module has about $5\times10^{12}$
possible choices. During searching, the models can be trained well
from scratch with a large batch size without using the pre-trained
ImageNet model. 

\subsection{Feature Fusion Search Module}

As mentioned in DetNAS \cite{chen2019detnas}, neurons in the later
stage of the backbone strongly respond to entire objects while other
neurons are more likely to be activated by local textures and patterns.
In the lane detection context, features in the later stage can capture
long-range coherent lane information while the features in the early
stage contain more accurate short-range curve information by its local
patterns. In order to fuse different information across multi-level
features, we propose a feature fusion search module to find a better
fusion of the high-level and low-level features. We also allow predictions
on different feature maps since the
detailed lane shapes are usually captured by a large feature map.
We consider the following search space for a lane-sensitive model:

Let $F_{1,...,t}$ denote the output feature maps from different stages
of the backbone ($t$ can be 3 or 4 depending on the choice of the
backbone). From $F_{1}$ to $F_{t}$, the spatial size is gradually
down-sampled with factor 2. Our feature fusion search module consists
of $M$ fusion layers $\left\{ O_{i}\right\} $. For each fusion layer
$O_{i}$, we pick two output feature levels with \{$F_{1}$, ... ,
$F_{4}$\} as input features and one target output resolution. The
two input features $F_{i}$ first go through one 1x1 convolution layer
to become one output channels $c$, then both features will do up-sampling/down-sampling
to the target output resolution and be concatenated together. The
output of each $O_{i}$ will go through another 1x1 convolution with
output channels $c$ and to concatenate to the final output. Thus
our search space is flexible enough to fuse different features and
select the output feature layer to feed into the heads. For each level
of the feature map, we also decide whether a prediction head should
be added (at least one). The total search space of the feature fusion
search module is relatively small (about $10^{3}$ possible choices).

\subsection{Adaptive Point Blending Search Module\label{subsec:Anchor-proposal-based-Lane}}

\begin{figure}[tb]

\begin{centering}
\includegraphics[height=3.2cm]{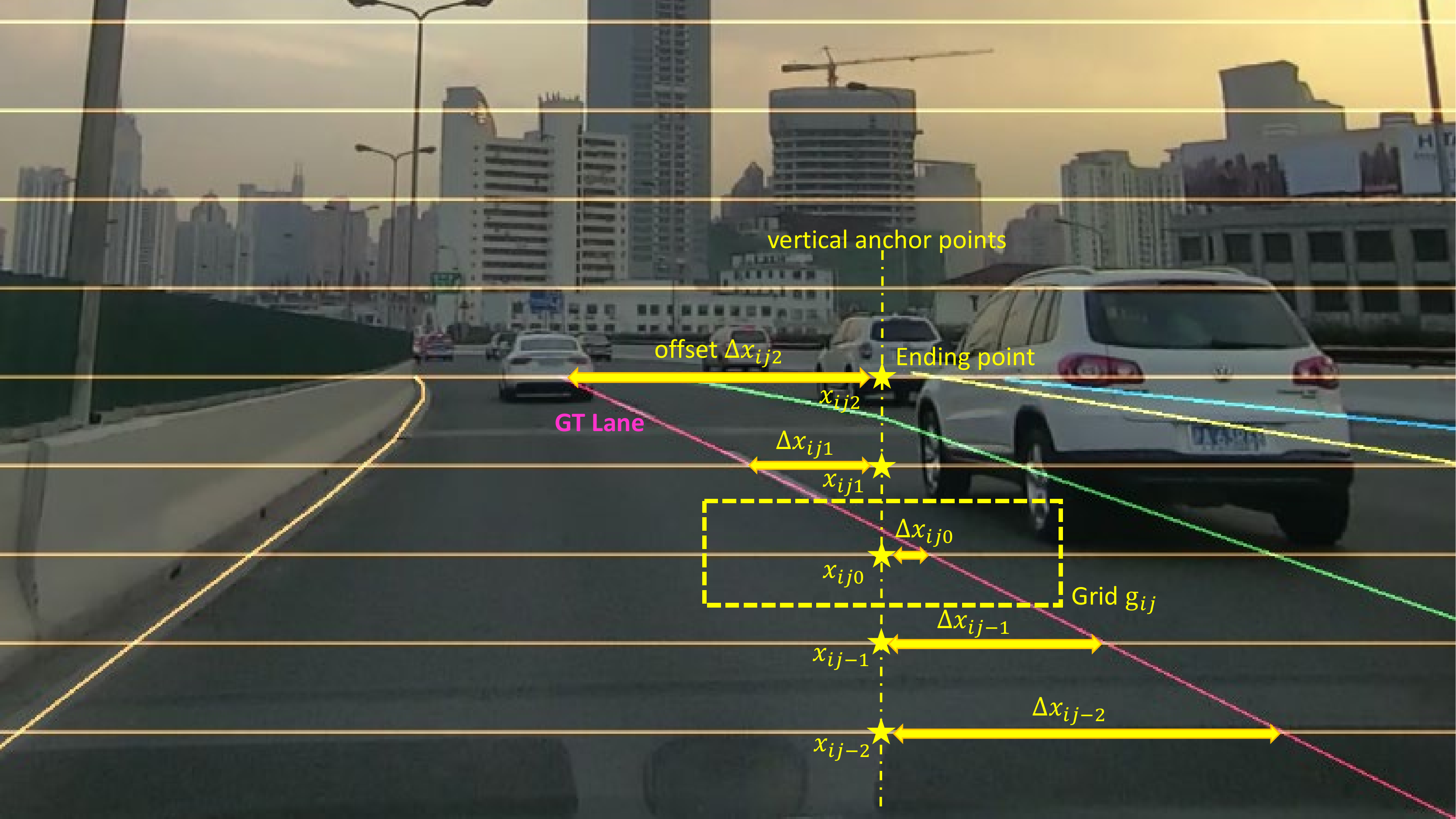}\includegraphics[height=3.2cm]{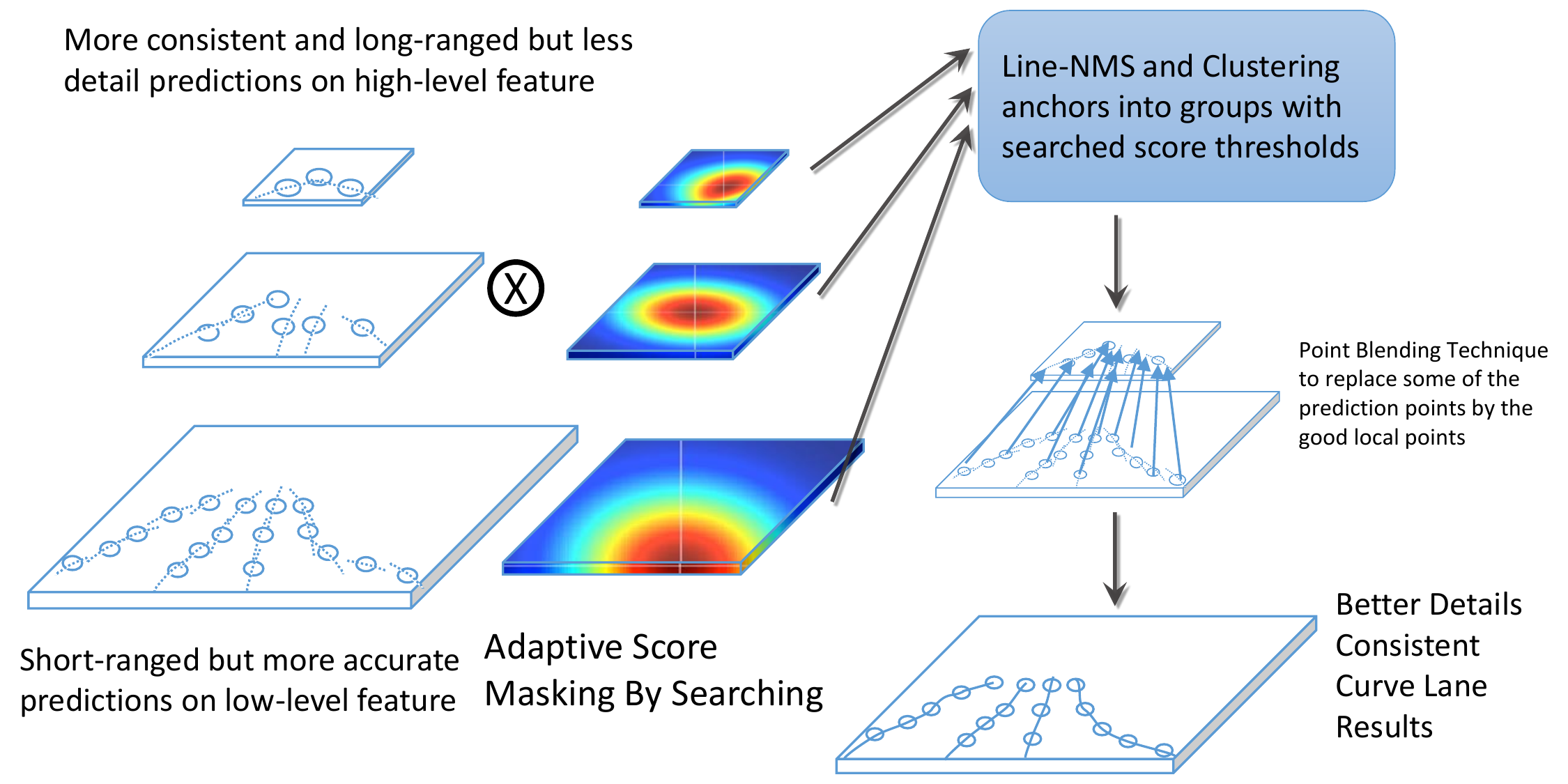}
\par\end{centering}

\caption{\label{fig:The-anchor-proposal-based}(Left) For each prediction head,
we predict the offsets $\varDelta x_{ijz}$ for each grid $g_{ij}$
and their corresponding scores. The offset $\varDelta x_{ijz}$ is
the horizontal distance between the ground truth and a pre-defined
vertical anchor $x_{ijz}$. Each lane prediction can be recovered
by the offset and the point anchors positions. (Right) We propose
an adaptive Point Blending Search Module for a lane-sensitive detection.
Adaptive score masking allows different regions of interest in multi-level
prediction. A point blending technique is used to replace some
of the prediction points in high confidence lines by the accurate
local points. The remote and detailed curve shape can be amended by
the local points.}

\end{figure}

Inspired by PointLaneNet \cite{chen2019pointlanenet}, each head proposes
many anchors on its feature map and predicts their corresponding offsets
to generate line proposals. A lane line can be determined in the image
by line points and one ending point. We first divide each feature
map into a $w_{f}$ \texttimes{} $h_{f}$ grid $G$. If the center
of the grid $g_{ij}$ is near to a ground truth lane, $g_{ij}$ is
responsible for detecting that lane. Since a lane will go across several
grids, multiple grids can be assigned to that lane and their confidence
scores $s_{ij}$ reflect how confident the grid contains a part of
the lane. For each $g_{ij}$, the model will predict a set of offsets
$\varDelta x_{ijz}$ and one ending point position, where $\varDelta x_{ijz}$
is the horizontal distance between the ground truth lane and a pre-defined
vertical anchor points $x_{ijz}$ as shown in Figure \ref{fig:The-anchor-proposal-based}.
With the predicted $\varDelta x_{ijz}$ and the ending point position,
each grid $g_{ij}$ can forecast one potential lane $l_{ij}$.\footnote{The description of the loss function can be found in the Appendix.}
 Post-processing is required to summarize and filtering all line
proposals and generate final results.

In PointLaneNet \cite{chen2019pointlanenet}, a Line-NMS is adopted
on one feature map to filter out lower confidence and a non-maximum
suppression (NMS) algorithm is used to filter out occluded lanes according
to their confidence score.\footnote{The detailed Line-NMS algorithm can be found in the Appendix.}
However, it cannot fit in our multi-level situation. First, predictions
from different levels of features cannot be treated equally. For example,
predictions on the low-level feature map are more accurate in a short-range
while predictions on high-level feature map are more consistent in
a long-range but losing a lot of details. Moreover, we found that
each grid can only predict the offsets precisely around its center
and the $\varDelta x_{ijz}$ far away from the anchor is inaccurate.
Using plain Line-NMS is not sensitive enough to capture the curve
or remote part of the lanes.

An adaptive Point Blending Search Module is proposed to unify the
post-processing into the NAS framework for lane-sensitive detection
as shown in Figure \ref{fig:The-anchor-proposal-based} (Right). In
order to allow a different emphasize regions in multi-level prediction,
we use an adaptive score masking $m_{f}$ on the original score prediction
for each feature map $f$. Let $c_{x}$ and $c_{y}$ denote the center
position of each grid $g_{ij}$ on certain feature map $f$. We consider
a very simple score masking for each map as follow:
\begin{equation}
logit(m_{f})=\alpha_{1f}(c_{y})+\beta_{1f}+\alpha_{2f}\left[(c_{x}-u_{xf})^{2}+(c_{y}-u_{yf})^{2}\right]^{\frac{1}{2}}.\label{eq:mask}
\end{equation}
There are two main terms in the Eq. (\ref{eq:mask}): one term linearly
related to the vertical position of each prediction, and another term
is related to the distance from $[u_{xf},u_{yf}]$. We use the above
masking because we conjecture that low-level features may perform
better in the remote part of the lane (near the center of the image)
and such formulation allows flexible masking across feature maps. 

Note that each grid with a good confidence score has precise local
information about the lane near the center of the grid. Filtering
out all other occluded lanes and only using the lane with the highest
confidence score in Line-NMS might not capture the long range curvature
for the remote part, since a low score may be assigned. Thus, we further
use a point blending technique for a lane-sensitive prediction. After
modification of the original confidence score on each feature map,
we first filter out those low score lanes by a suitable threshold
and apply NMS to group the remaining lanes into several groups according
to their mutual distance. In each group of lines, we iteratively swap
the good local points in the lower score anchors with those remote
points in the highest score anchors. For each high confidence anchor,
some of its points are then replaced by the good local points to become
the final prediction. The remote parts of lanes and curve shape can
be amended by the local points and the details of lanes are better.
The computation overhead is very small by adding a loop within each
group of limited lines. The detailed algorithm can be found in the
Appendix.

Thus, $\alpha_{1f}$, $\beta_{1f}$, $\alpha_{2f}$ , $[u_{xf},u_{yf}]$,
the score thresholds and the mutual distance thresholds in NMS form
the final search space of this module. 

\subsection{Unified Multi-objective Search}

We consider a simple but effective multi-objective search algorithm
that can generate a Pareto front with the optimal trade-off between
accuracy and different computation constraints. Non-dominate sorting
is used to determine whether one model dominates another model in
terms of both FLOPS and accuracy. During the search, we sample a candidate
architecture by mutating the best architecture along the current Pareto
front. We considering following mutation: for the backbone search
space, we randomly swap the position of downsampling and double-channel
to their neighboring position; for the feature fusion search space,
we randomly change the input feature level of each fusion layer; for
the adaptive point blending search module, we disturb the best hyper-parameters.
Note that the search of the post-processing parameters does not involve
training thus the mutation can be more frequent on this module. Our
algorithm can be run on multiple parallel computation nodes and can
lift the Pareto front simultaneously. As a result, the search algorithm
will automatically allocate computation with reasonable receptive
fields and spatial resolution towards an effecient and lane-sensitive
detection. 

\begin{figure}[tb]

\begin{centering}
\includegraphics[width=0.2\columnwidth]{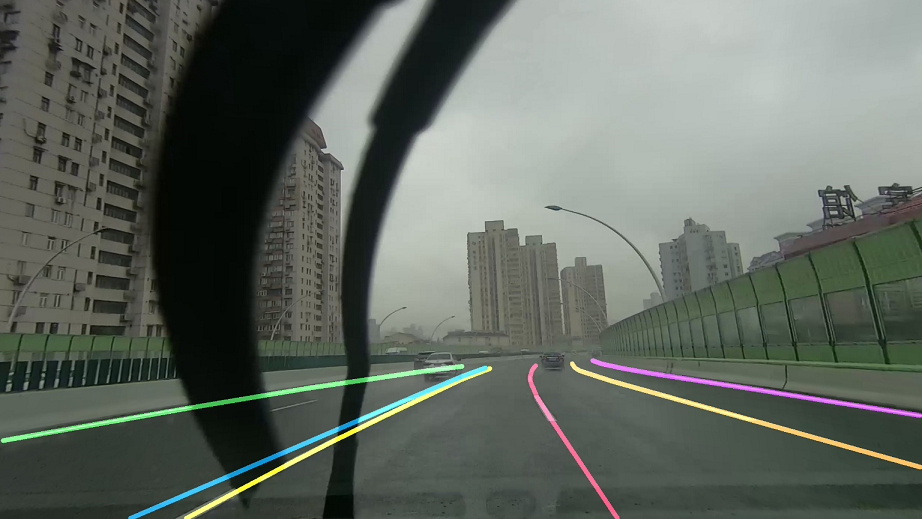}\includegraphics[width=0.2\columnwidth]{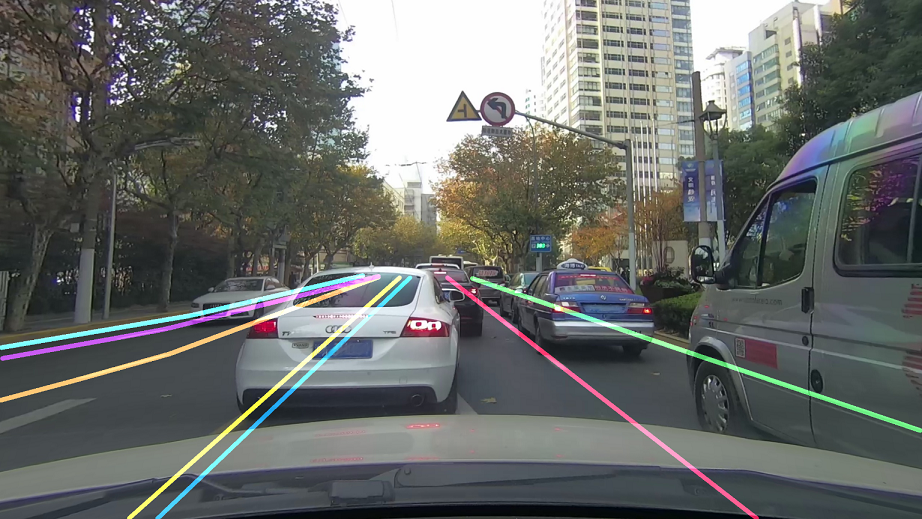}\includegraphics[width=0.2\columnwidth]{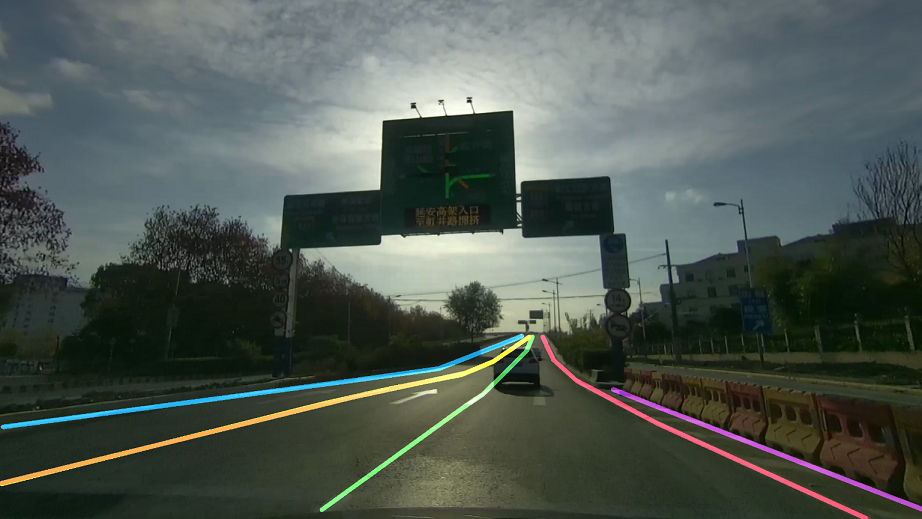}\includegraphics[width=0.2\columnwidth]{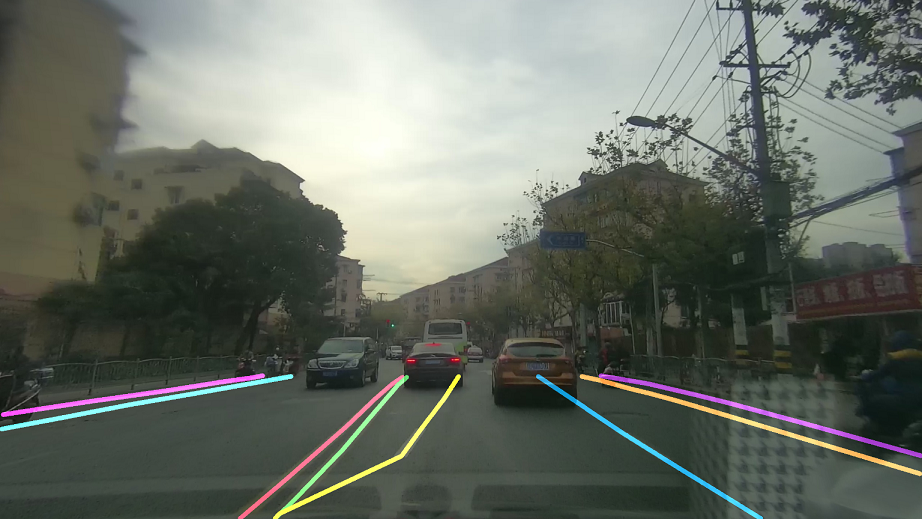}\includegraphics[width=0.2\columnwidth]{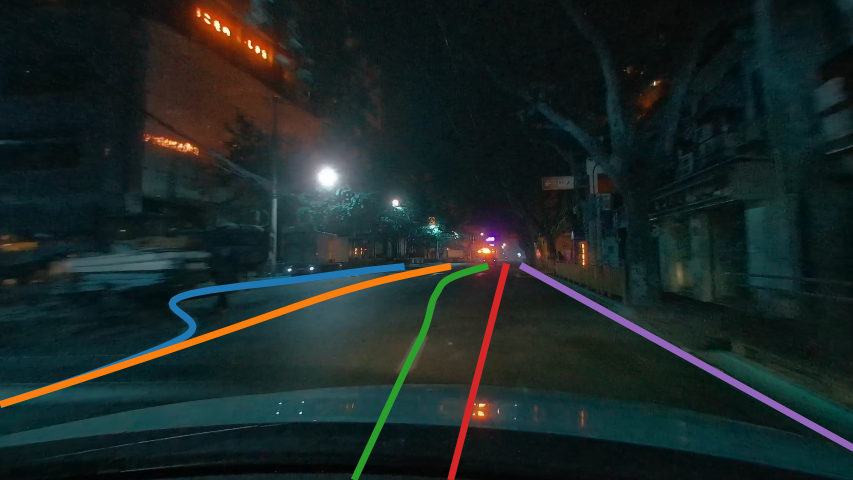}
\par\end{centering}
\begin{centering}
\includegraphics[width=0.2\columnwidth]{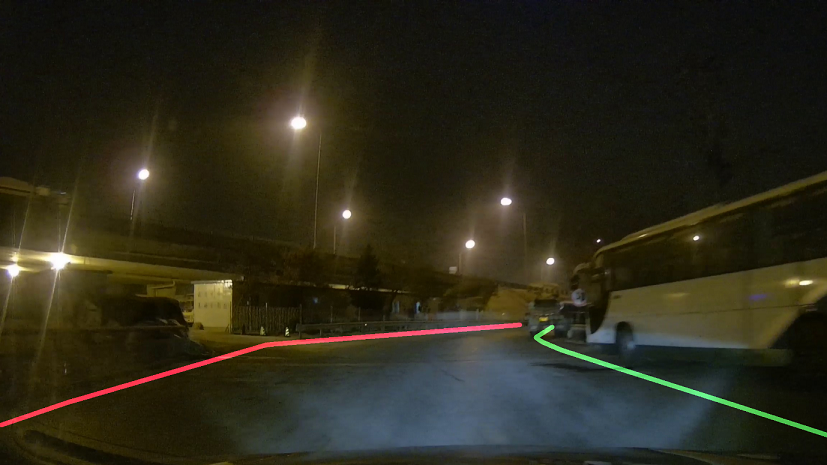}\includegraphics[width=0.2\columnwidth]{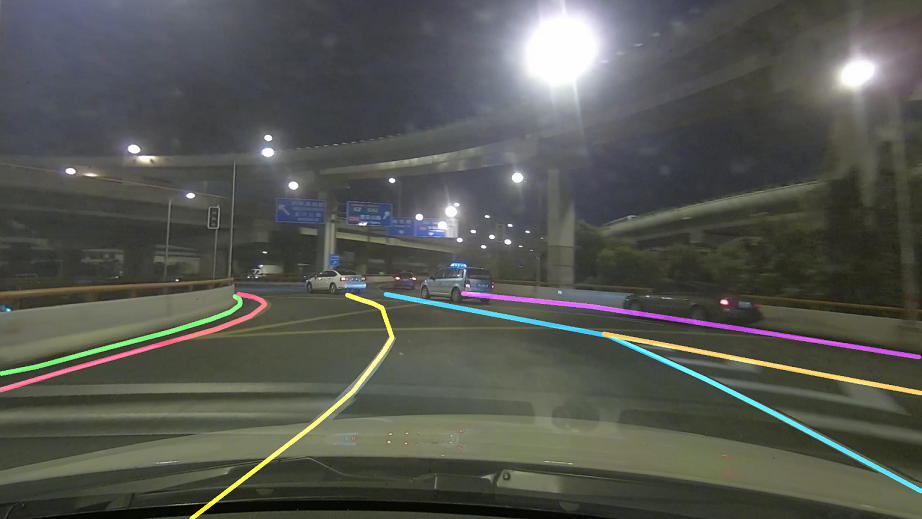}\includegraphics[width=0.2\columnwidth]{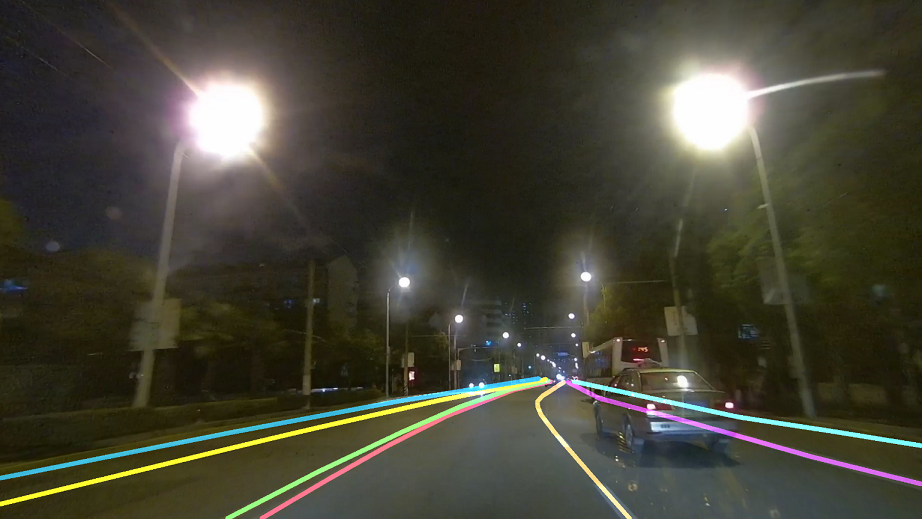}\includegraphics[width=0.2\columnwidth]{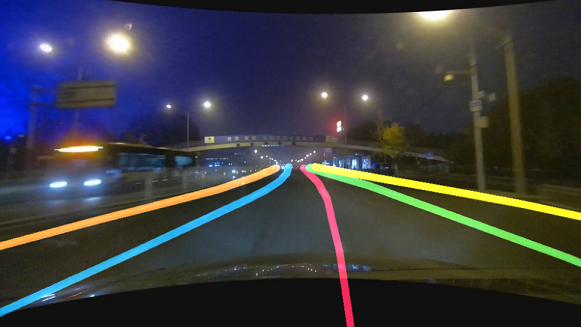}\includegraphics[width=0.2\columnwidth]{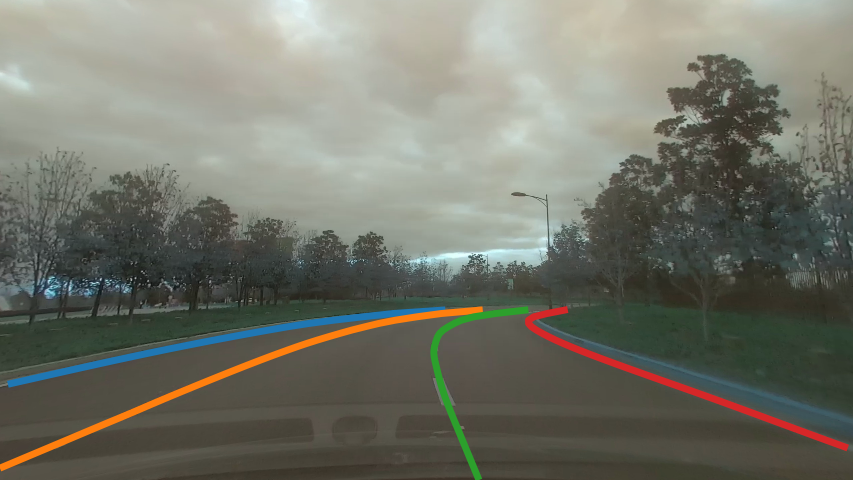}
\par\end{centering}
\begin{centering}
\includegraphics[width=0.2\columnwidth]{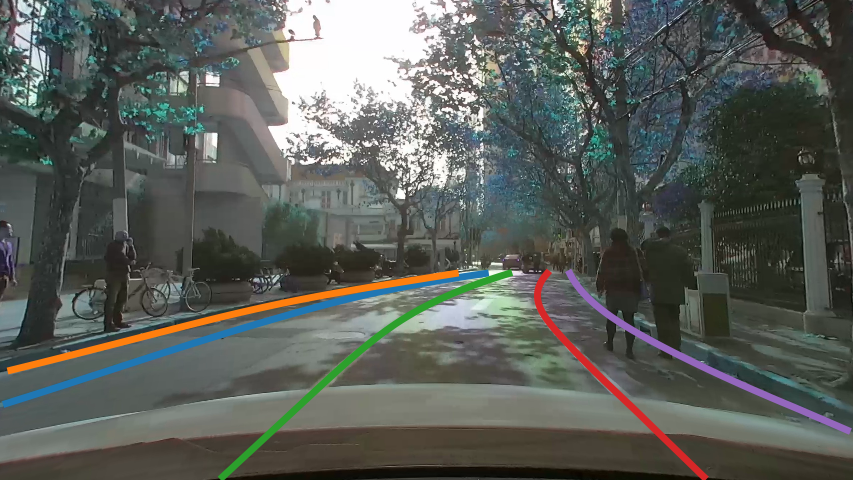}\includegraphics[width=0.2\columnwidth]{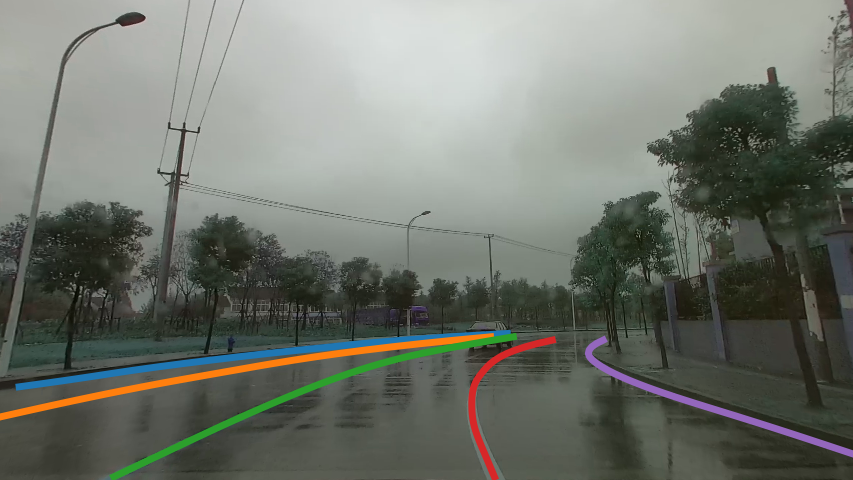}\includegraphics[width=0.2\columnwidth]{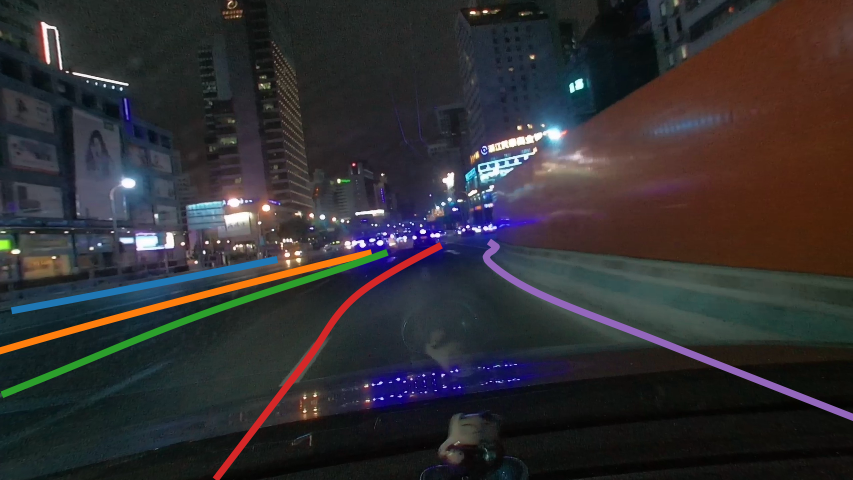}\includegraphics[width=0.2\columnwidth]{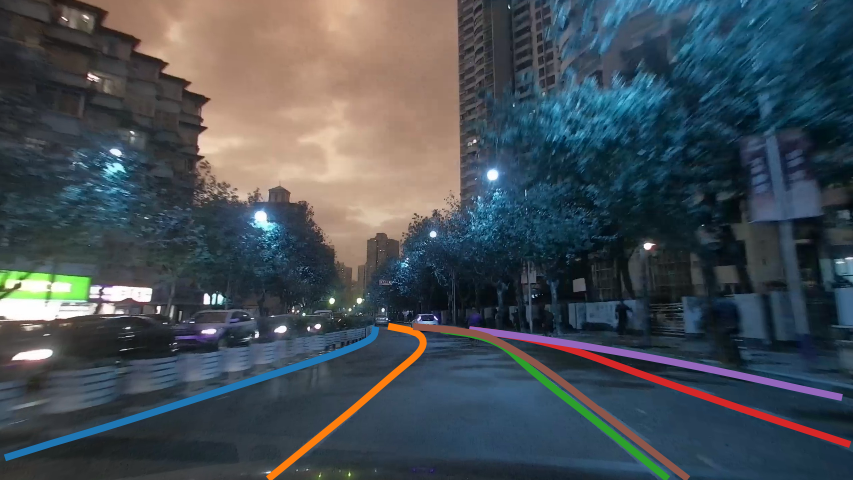}\includegraphics[width=0.2\columnwidth]{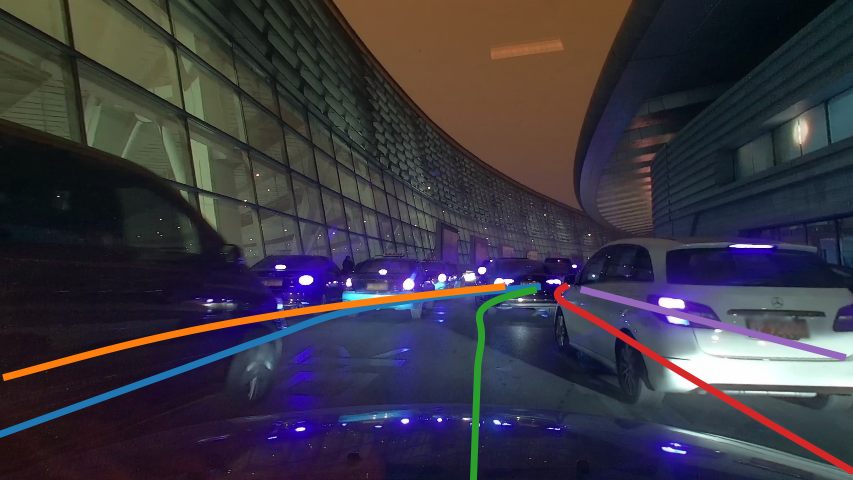}
\par\end{centering}

\caption{\label{fig:Examples-of-CurveLanes}Examples of our new released CurveLanes
dataset. All images are carefully selected so that each image contains
at least one curve lane. It is the largest lane detection dataset
so far and establishes a more challenging benchmark for the community.
More difficult scenarios such as S-curves and Y-lanes can be found
in this dataset.}
\end{figure}

\section{Experiments}

\subsection{New CurveLanes Benchmark}

\begin{table}[tb]
\caption{\label{tab:Comparsion-of-three-dataset}Comparison of the three largest
lane detection datasets and our new CurveLanes. Our new CurveLanes
benchmark has substantially more images, bigger resolution, more average
number of lanes and more curves lanes.}

\begin{centering}
\tabcolsep 0.02in{\footnotesize{}}%
\begin{tabular}{cccccc}
\hline 
{\scriptsize{}Datasets} & {\scriptsize{}Total amount of images} & {\scriptsize{}Resolution} & {\scriptsize{}Road type} & {\scriptsize{}\# Lane\textgreater 5} & {\scriptsize{}Curves}\tabularnewline
\hline 
{\scriptsize{}TuSimple \cite{TuSimple}} & {\scriptsize{}13.2K} & {\scriptsize{}1280x720} & {\scriptsize{}Highway} & {\scriptsize{}$\times$} & {\scriptsize{}\textasciitilde 30\%}\tabularnewline
{\scriptsize{}CULane \cite{pan2018spatial}} & {\scriptsize{}133.2K} & {\scriptsize{}1640x590} & {\scriptsize{}Urban \& Highway} & {\scriptsize{}$\times$} & {\scriptsize{}\textasciitilde 2\%}\tabularnewline
{\scriptsize{}BDD100K \cite{yu2018bdd100k}} & {\scriptsize{}100K} & {\scriptsize{}1280x720} & {\scriptsize{}Urban \& Highway} & \textbf{\scriptsize{}$\checked$} & {\scriptsize{}\textasciitilde 10\%}\tabularnewline
{\scriptsize{}Our CurveLanes} & \textbf{\scriptsize{}150K} & \textbf{\scriptsize{}2650x1440} & {\scriptsize{}Urban \& Highway} & \textbf{\scriptsize{}$\checked$} & {\scriptsize{}\textgreater}\textbf{\scriptsize{}90\%}\tabularnewline
\hline 
\end{tabular}{\footnotesize\par}
\par\end{centering}

\end{table}

\begin{figure}[tb]

\begin{centering}
\includegraphics[scale=0.3]{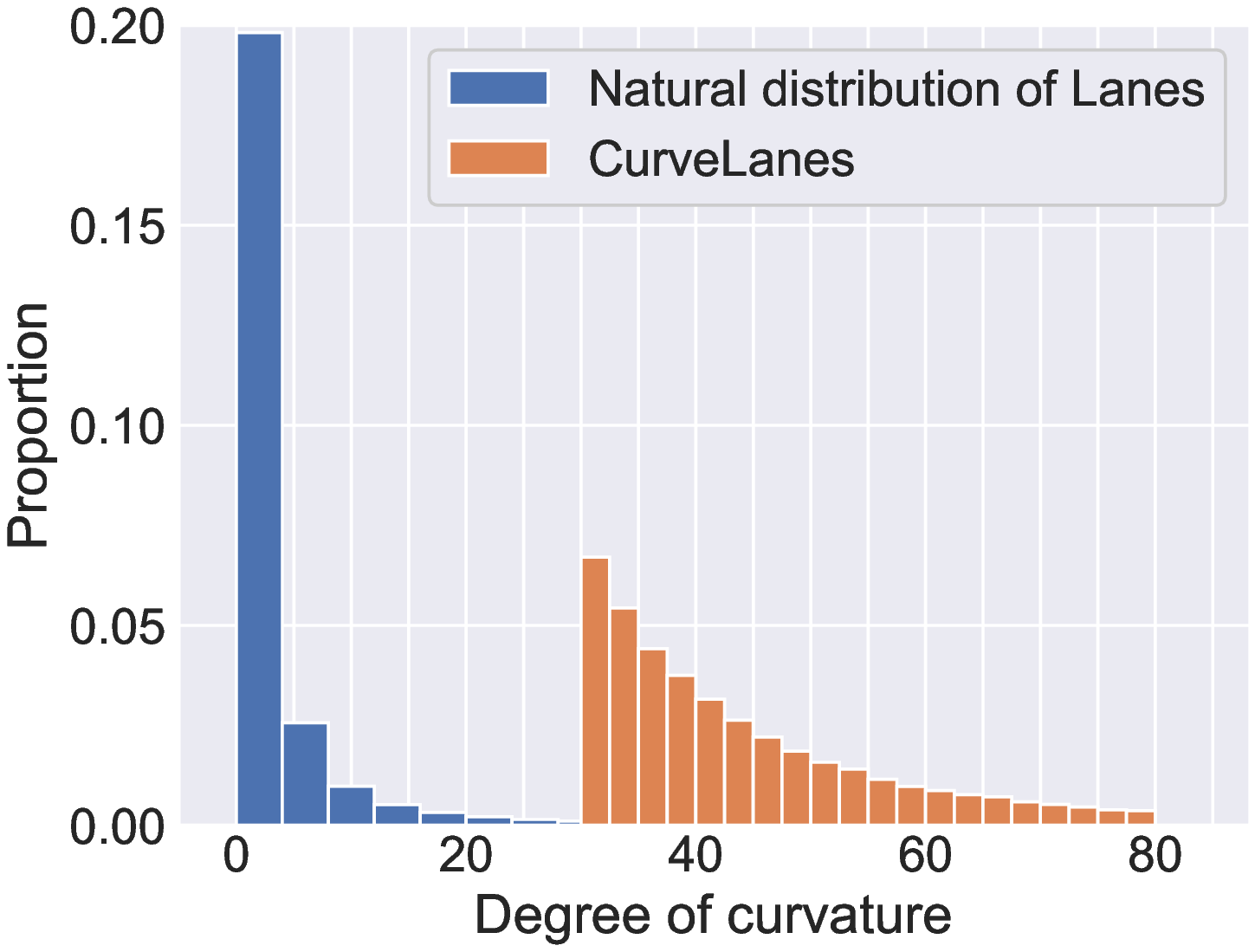}\includegraphics[scale=0.3]{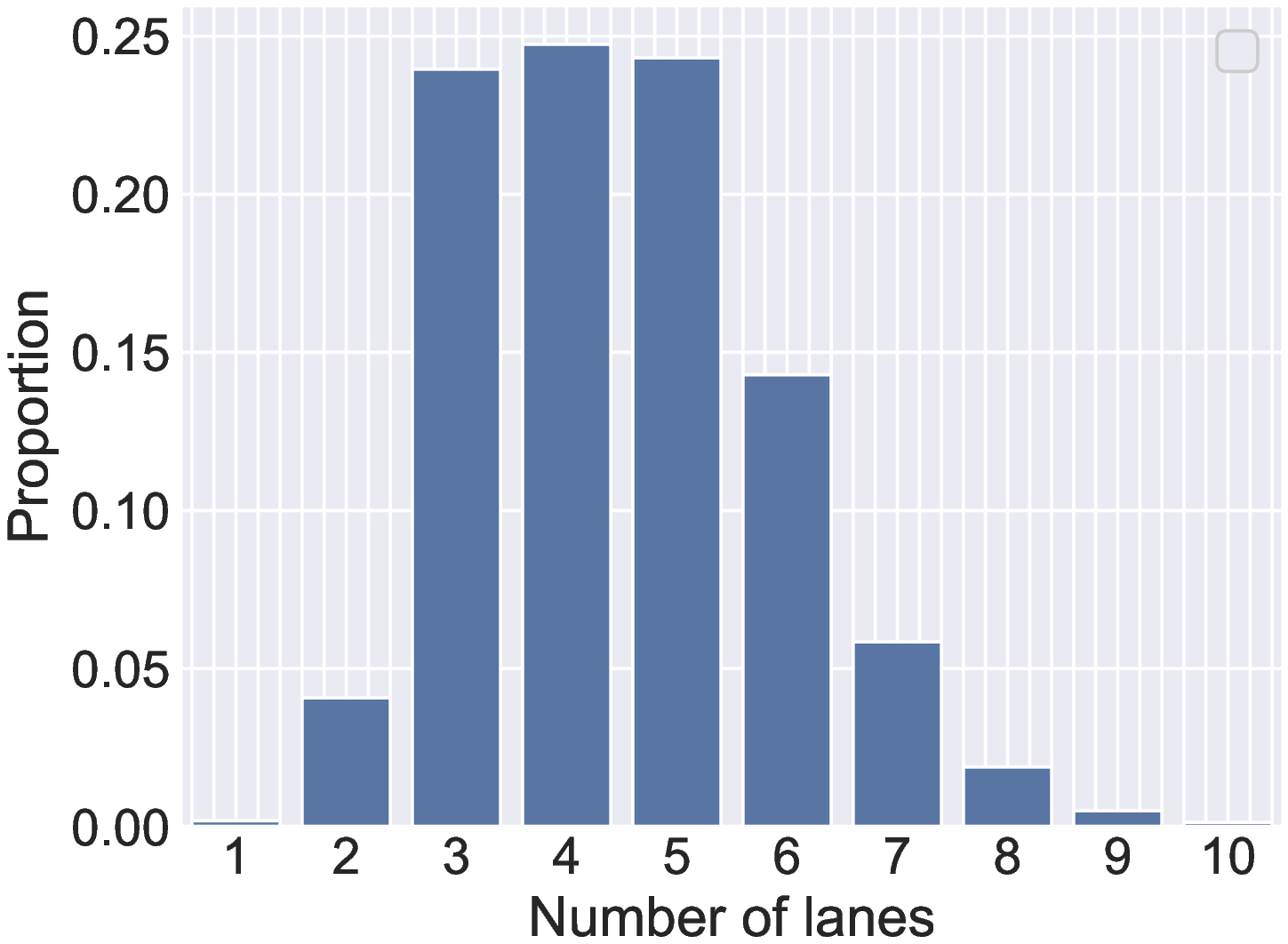}
\par\end{centering}

\caption{\label{fig:Examples-of-CurveLanes-1}(Left) Comparison of the distribution
of the degree of curvature between common datasets and our CurveLanes.
Ours has more proportion of various curvatures comparing to the natural
distribution of lanes. (Right) The histogram of the average number
of lanes per image in our CurveLanes. CurveLanes also has more number
of lanes per image than CULane (\textless 5) and TuSimple (\textless 5)
thus more challenging.}

\end{figure}

We have released a new dataset named CurveLanes consisting of 150K
lanes images with 650K carefully annotated lane labels for bench-marking
difficult scenarios such as curves and multi-lanes in traffic lane
detection. Table \ref{tab:Comparsion-of-three-dataset} shows a comparison
between the existing lane detection datasets TuSimple \cite{TuSimple},
CULane \cite{pan2018spatial} and BDD \cite{yu2018bdd100k}. It can
be found that our new CurveLanes benchmark has more images and a higher
resolution. Moreover, our dataset has more lanes per image and more
curves lanes. Thus, the new benchmark is suitable to compare the performance
in the difficult situation for the community.

Figure \ref{fig:Examples-of-CurveLanes} shows some typical examples
of the CurveLanes. More difficult cases such as S-curves, Y-lanes
can be found in this dataset. Figure \ref{fig:Examples-of-CurveLanes-1}
further shows the comparison of the distribution of the degree of
curvature between common dataset and CurveLanes. The new CurveLanes has 
more turns and difficult curves CurveLanes
also has more muti-lanes scenes thus more difficult. The whole dataset
150K is divided into: train:100K, val: 20K and testing
30K.

\begin{table}[tb]

\caption{\label{tab:Performance-on-CULane}Comparison of F1-measure of the
state-of-the-art models on \textbf{CULane} test set. CurveLane-S,
CurveLane-M, and CurveLane-L are the searched architectures of our
method. Our method outperforms the SOTA models by a large margin with
a small computational overhead.}

\begin{centering}
{\scriptsize{}\tabcolsep 0.00005in}%
\begin{tabular}{c|cccc|ccc}
\hline 
{\scriptsize{}Methods} & {\scriptsize{}SCNN\cite{pan2018spatial}} & {\scriptsize{}SAD\cite{hou2019learning}} & {\scriptsize{}SAD} & {\scriptsize{}PointLane\cite{chen2019pointlanenet}} & \textbf{\scriptsize{}CurveLane-S} & \textbf{\scriptsize{}CurveLane-M} & \textbf{\scriptsize{}CurveLane-L}\tabularnewline
\hline 
{\scriptsize{}Backbone} & {\scriptsize{}SCNN} & {\scriptsize{}ENet } & {\scriptsize{}R101} & {\scriptsize{}R101 } & {\scriptsize{}Searched} & {\scriptsize{}Searched} & {\scriptsize{}Searched}\tabularnewline
\hline 
{\scriptsize{}Normal} & {\scriptsize{}90.6} & {\scriptsize{}90.1} & {\scriptsize{}90.7} & {\scriptsize{}88.0} & {\scriptsize{}88.3} & {\scriptsize{}90.2} & {\scriptsize{}90.7}\tabularnewline
{\scriptsize{}Crowded} & {\scriptsize{}69.7} & {\scriptsize{}68.8} & {\scriptsize{}70.0} & {\scriptsize{}68.1} & {\scriptsize{}68.6} & {\scriptsize{}70.5} & {\scriptsize{}72.3}\tabularnewline
{\scriptsize{}Dazzle light} & {\scriptsize{}58.5} & {\scriptsize{}60.2} & {\scriptsize{}59.9} & {\scriptsize{}61.5} & {\scriptsize{}63.2} & {\scriptsize{}65.9} & {\scriptsize{}67.7}\tabularnewline
{\scriptsize{}Shadow} & {\scriptsize{}66.9} & {\scriptsize{}65.9} & {\scriptsize{}67.0} & {\scriptsize{}63.3} & {\scriptsize{}68.0} & {\scriptsize{}69.3} & {\scriptsize{}70.1}\tabularnewline
{\scriptsize{}No line} & {\scriptsize{}43.4} & {\scriptsize{}41.6} & {\scriptsize{}43.5} & {\scriptsize{}44.0} & {\scriptsize{}47.9} & {\scriptsize{}48.8} & {\scriptsize{}49.4}\tabularnewline
{\scriptsize{}Arrow} & {\scriptsize{}84.1} & {\scriptsize{}84.0} & {\scriptsize{}84.4} & {\scriptsize{}80.9} & {\scriptsize{}82.5} & {\scriptsize{}85.7} & {\scriptsize{}85.8}\tabularnewline
{\scriptsize{}Curve} & {\scriptsize{}64.4} & {\scriptsize{}65.7} & {\scriptsize{}65.7} & {\scriptsize{}65.2} & {\scriptsize{}66.0} & {\scriptsize{}67.5} & {\scriptsize{}68.4}\tabularnewline
{\scriptsize{}Night} & {\scriptsize{}66.1} & {\scriptsize{}66.0} & {\scriptsize{}66.3} & {\scriptsize{}63.2} & {\scriptsize{}66.2} & {\scriptsize{}68.2} & {\scriptsize{}68.9}\tabularnewline
{\scriptsize{}Crossroad} & {\scriptsize{}1990} & {\scriptsize{}1998} & {\scriptsize{}2052} & {\scriptsize{}1640} & {\scriptsize{}2817} & {\scriptsize{}2359} & {\scriptsize{}1746}\tabularnewline
\hline 
{\scriptsize{}FLOPS (G)} & {\scriptsize{}328.4} & {\scriptsize{}3.9} & {\scriptsize{}162.2} & {\scriptsize{}25.1} & {\scriptsize{}9.0} & {\scriptsize{}35.7} & {\scriptsize{}86.5}\tabularnewline
\hline 
\textbf{\scriptsize{}Total} & {\scriptsize{}71.6} & {\scriptsize{}70.8} & {\scriptsize{}71.8} & {\scriptsize{}70.2} & \textbf{\scriptsize{}71.4} & \textbf{\scriptsize{}73.5} & \textbf{\scriptsize{}74.8}\tabularnewline
\hline 
\end{tabular}{\scriptsize\par}
\par\end{centering}

\end{table}

\subsection{Other Datasets and Evaluation Metrics}

We conduct neural architecture search on two large lane detection
datasets: the CULane \cite{pan2018spatial}, and the new CurveLanes
dataset. We also transfer the searched architectures to the TuSimple
\cite{TuSimple} and test the generalization power of the proposed
approach.\textbf{ CULane} \cite{pan2018spatial} is a large scale
dataset on traffic lane detection which is collected by cameras in
Beijing, China. The CULane dataset includes 88,880 training images,
9675 verification images, and 34,680 test images. The test dataset
is divided into 1 normal and 8 challenging categories.\textbf{ TuSimple
}\cite{TuSimple} is created by TuSimple specifically focuses on real
highway scenarios. It includes 3626 training images and 2782 test
images. 

\textbf{Evaluation metrics.} Evaluation metrics is important since
it is also the target of our architecture search. We follow the literature
\cite{pan2018spatial} and use the corresponding evaluation metrics
for each particular dataset. 1) CULane and CurveLanes. Following the
official implementation of the evaluation \cite{pan2018spatial},
we compute the intersection-over-union (IoU) between GT labels and
predictions, where each lane has 30 pixel width. Predictions whose
IoUs are larger than 0.5 are considered as true positives (TP). The
F1 measure is used as the evaluation metric: $F_{1}=\frac{2\times Precision\times Recall}{Precision+Recall}$,
where $Precision=\frac{TP}{TP+FP}$ and $Recall=\frac{TP}{TP+FN}$.
2) TuSimple. We also use the official metric as the evaluation metrics:
$Accuracy=\frac{N_{pred}}{N_{GT}}$, where where $N_{pred}$ is the
number of correctly predicted lane points and $N_{GT}$ is the number
of ground-truth lane points.

\begin{table}[tb]

\caption{\label{tab:CurveLanes}Comparison of different algorithms on the new
dataset \textbf{CurveLanes}. CurveLane-S, CurveLane-M, and CurveLane-L
are the searched architectures of our method. The SOTA methods such
as SCNN and SAD suffer substantial performance drop (20\%\textasciitilde 30\%
F1 score).}

\begin{centering}
\tabcolsep 0.03in{\footnotesize{}}%
\begin{tabular}{ccccc}
\hline 
{\footnotesize{}Method} & {\footnotesize{}F1} & {\footnotesize{}Precision} & {\footnotesize{}Recall} & {\footnotesize{}FLOPS(G)}\tabularnewline
\hline 
{\footnotesize{}SCNN \cite{pan2018spatial}} & {\footnotesize{}65.02\%} & {\footnotesize{}76.13\%} & {\footnotesize{}56.74\%} & {\footnotesize{}328.4}\tabularnewline
{\footnotesize{}Enet-SAD \cite{hou2019learning}} & {\footnotesize{}50.31\%} & {\footnotesize{}63.6\%} & {\footnotesize{}41.6\%} & {\footnotesize{}3.9}\tabularnewline
{\footnotesize{}PointLaneNet \cite{chen2019pointlanenet}} & {\footnotesize{}78.47\%} & {\footnotesize{}86.33\%} & {\footnotesize{}72.91\%} & {\footnotesize{}14.8}\tabularnewline
\hline 
\textbf{\footnotesize{}CurveLane-S} & {\footnotesize{}81.12\%} & {\footnotesize{}93.58\%} & {\footnotesize{}71.59\%} & {\footnotesize{}7.4}\tabularnewline
\textbf{\footnotesize{}CurveLane-M} & {\footnotesize{}81.80\%} & {\footnotesize{}93.49\%} & {\footnotesize{}72.71\%} & {\footnotesize{}11.6}\tabularnewline
\textbf{\footnotesize{}CurveLane-L} & \textbf{\footnotesize{}82.29\%} & {\footnotesize{}91.11\%} & {\footnotesize{}75.03\%} & {\footnotesize{}20.7}\tabularnewline
\hline 
\end{tabular}{\footnotesize\par}
\par\end{centering}

\end{table}

\textbf{NAS Implementation Details.} 
During the search, we directly trained the model without ImageNet
pretraining since the architecture of the backbone is changed. We
found that for a large dataset like CULane and CurveLanes(more than
80K training images), ImageNet pretraining is not necessary and the
resulting model converges well with only about 3\textasciitilde 5\%
accuracy loss. We use FLOPS to measure the computational complexity and construct the Pareto front.
During search, we use SGD with cosine decay learning
rate 0.04 to 0.0001, momentum $0.9$. We train each candidate for
12 epochs. Empirically, we found that training with 12 epochs can
well separate good models from bad models. We train and test the new
architecture in parallel on four computation nodes, and each has 8
Nvidia V100 GPU cards. The batch size is 256 and the input size is
$512\times288$. It takes about 1 hour to complete evaluating one
architecture for both datasets and it only takes 5 minutes to evaluate
one setting of post-processing parameters. The total search cost is
about 5000 GPU hours for one dataset. We set the number of blocks
from 10-45 in order to get a complete Pareto front with different
FLOPS. We set two random fusions ($M=2$) with 128 output channels
for the fusion search module.

\begin{figure}[tb]
\begin{centering}
\begin{sideways}{GT}\end{sideways} \includegraphics[width=0.18\columnwidth]{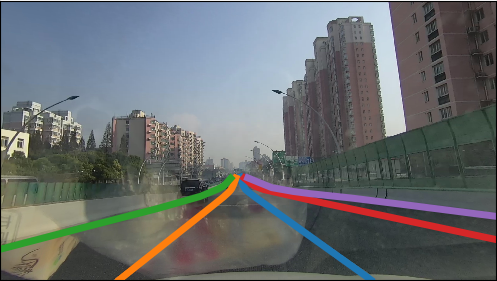}\includegraphics[width=0.18\columnwidth]{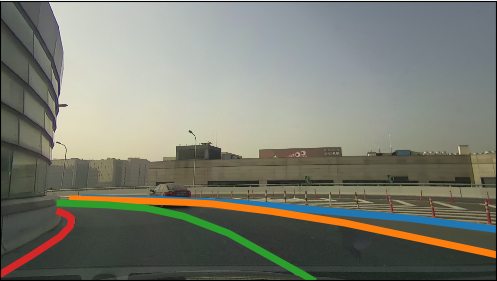}\includegraphics[width=0.18\columnwidth]{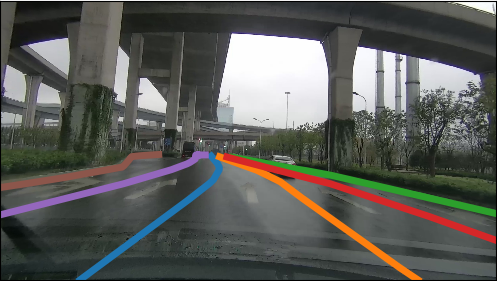}\includegraphics[width=0.18\columnwidth]{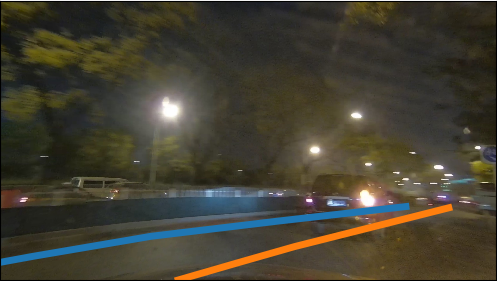}\includegraphics[width=0.18\columnwidth]{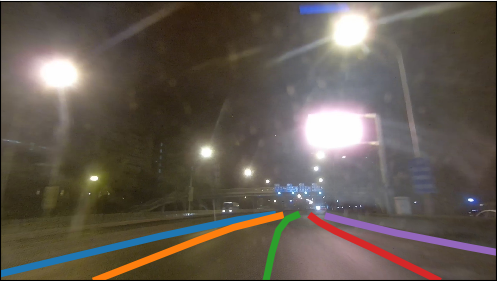}
\par\end{centering}
\begin{centering}
\begin{sideways}{SCNN}\end{sideways} \includegraphics[width=0.18\columnwidth]{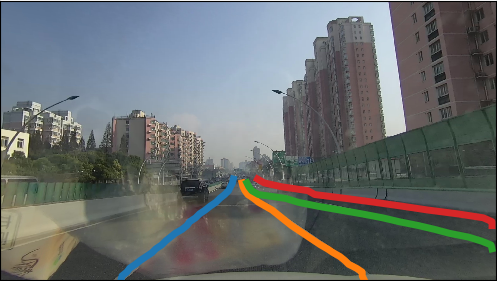}\includegraphics[width=0.18\columnwidth]{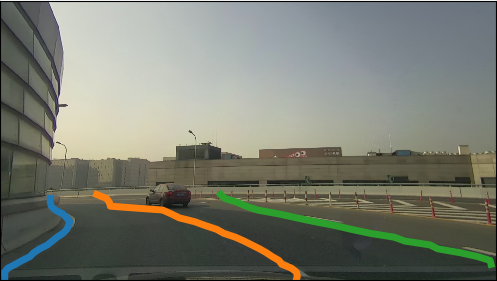}\includegraphics[width=0.18\columnwidth]{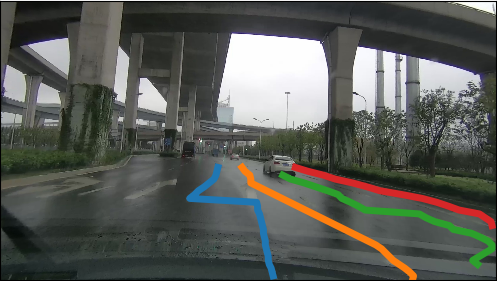}\includegraphics[width=0.18\columnwidth]{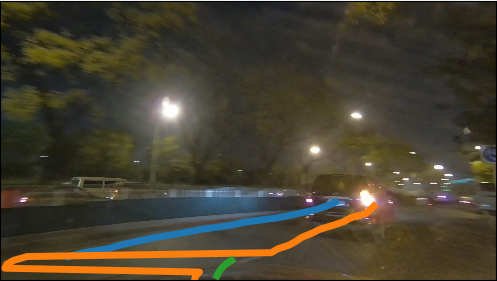}\includegraphics[width=0.18\columnwidth]{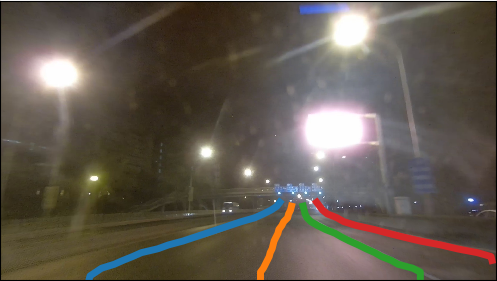}
\par\end{centering}
\begin{centering}
\begin{sideways}{SAD}\end{sideways} \includegraphics[width=0.18\columnwidth]{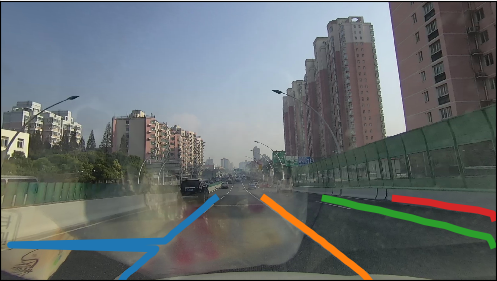}\includegraphics[width=0.18\columnwidth]{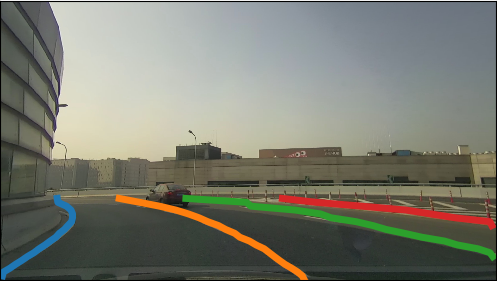}\includegraphics[width=0.18\columnwidth]{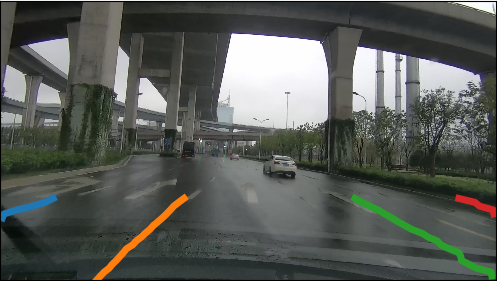}\includegraphics[width=0.18\columnwidth]{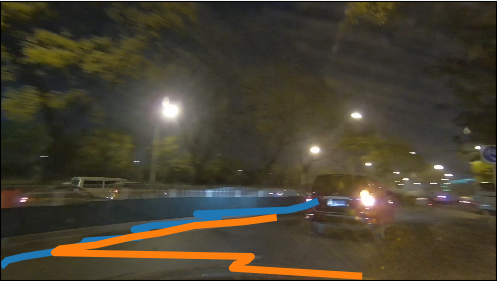}\includegraphics[width=0.18\columnwidth]{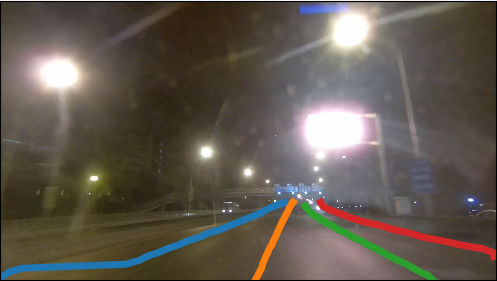}
\par\end{centering}
\begin{centering}
\begin{sideways}{Ours}\end{sideways} \includegraphics[width=0.18\columnwidth]{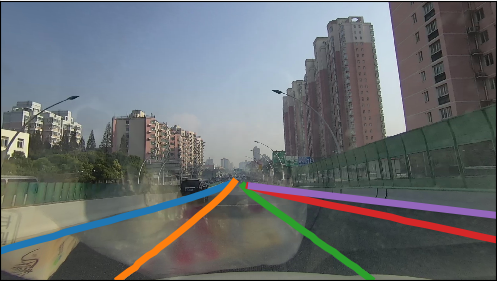}\includegraphics[width=0.18\columnwidth]{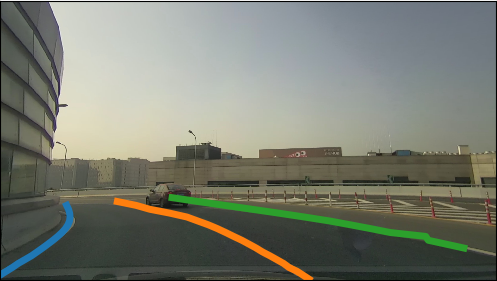}\includegraphics[width=0.18\columnwidth]{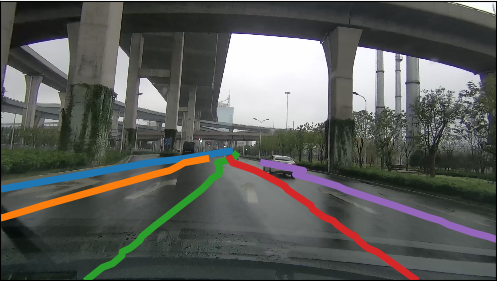}\includegraphics[width=0.18\columnwidth]{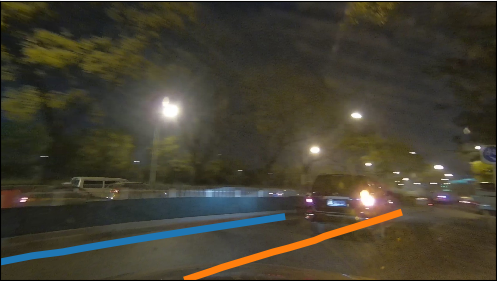}\includegraphics[width=0.18\columnwidth]{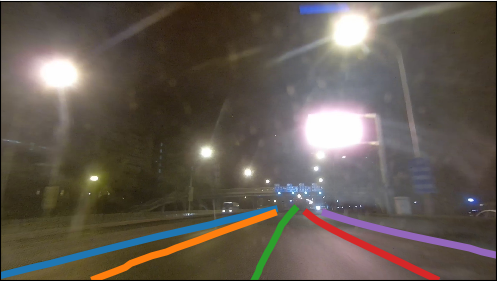}
\par\end{centering}

\caption{\label{fig:Results-of-CurveLanes}Qualitative result comparison on
CurveLanes. Our method CurveLane-L performs
better in the difficult scenarios such as curves, night and wet roads.}

\end{figure}

\textbf{}

\subsection{Lane Detection Results}

After identifying the optimal architecture on each dataset, we fully
train those models. We first pre-train those searched backbones on
ImageNet following common practice \cite{he2016deep} for fair comparison
with other methods. We train 50 epochs with $bs=256$, 40 epochs with
$bs=32$ and 30 epochs with $bs=$32 for TuSimple, CULane and CurveLanes,
respectively. SGD is used with initial learning rate 0.04 and a cosine
decay learning rate $0.04$ to $0.0001$, momentum $0.9$. The input
size of both training and testing is $512\times288$ for three datasets.
We consider three kinds of computational constraints thus the resulting
models are denoted as CurveLane-S, CurveLane-M, and CurveLane-L picked
from the Pareto front of each dataset.  A detailed description of
the hyper-parameter can be found in the supplementary materials. For
other methods, we report the accuracy numbers of \cite{pan2018spatial,hou2019learning}
directly from the original papers for TuSimple and CULane. For the
CurveLanes, we re-implement the official code from the \cite{pan2018spatial,hou2019learning}. For PointLaneNet\cite{chen2019pointlanenet},
we use ResNet101 as the backbone. 

\begin{figure}[tb]

\begin{centering}
{\tiny{}\includegraphics[width=0.45\columnwidth]{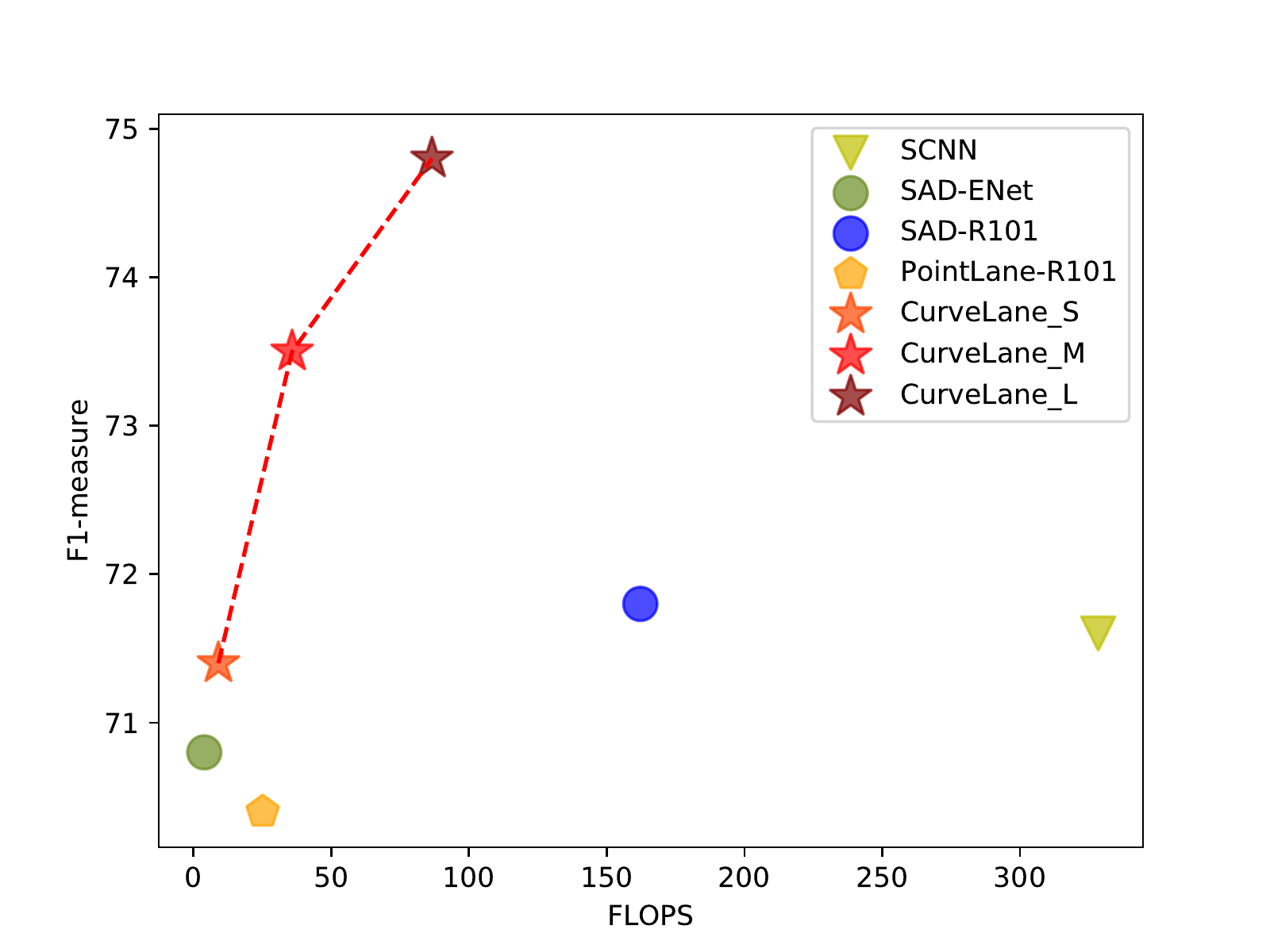}\includegraphics[width=0.45\columnwidth]{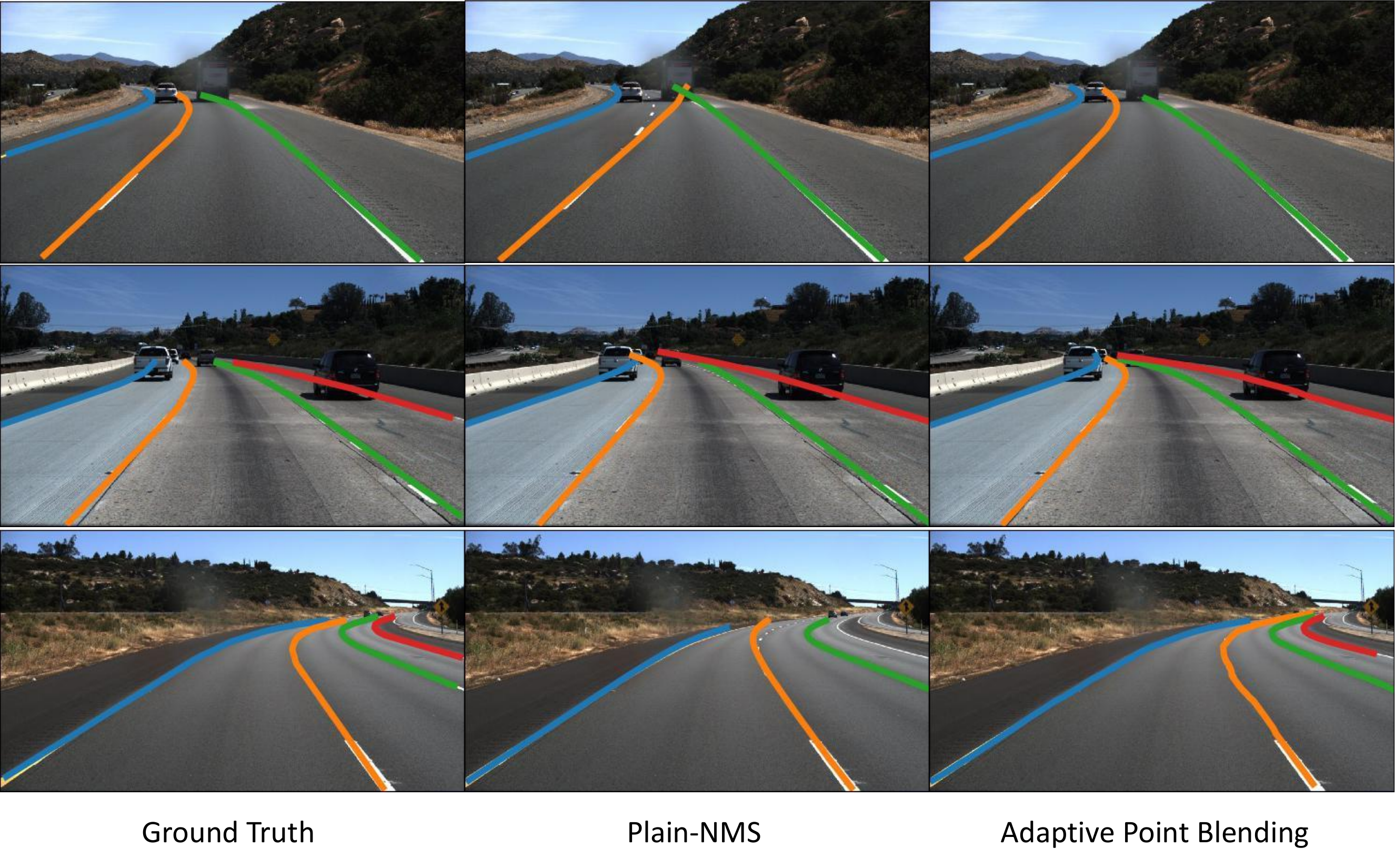}}{\tiny\par}
\par\end{centering}

\caption{\label{fig:Performance-of-the-NMS-SWAP}(Left) Comparison between
our methods and other SOTA methods on CULane. Although specifically
designed for curve lane detection, our method still dominates most
SOTA methods on the widely used CULane benchmark. (Right) Performance
of the post-processing algorithms on TuSimple dataset: Ground-truth,
Plain-NMS, and our Adaptive Point Blending. Our methods are more sensitive
to the curve lanes and remote part of the lanes.}

\end{figure}

\textbf{Comparison with the state-of-the-art on CULane.} Table \ref{tab:Performance-on-CULane}
shows the performance of the searched architectures on the CULane.
Our CurveLane-L achieves a new SOTA result on CULane dataset with
a 74.8 F1 score comparing to all the competing methods. Our CurveLane-M
model is 1.9 higher F1-core, and 9x fewer FLOPS than SCNN; 1.5 higher
F1-score, 4.5x fewer FLOPS than R101-SAD. Figure \ref{fig:Performance-of-the-NMS-SWAP}
(Left) shows a comparison between ours and other SOTA methods. Although
our method is specifically designed for curve lane detection, it dominates
most SOTA methods which proves the effectiveness of the proposed multi-objective
search framework. 

\textbf{Results on new CurveLanes.} The comparison is shown in Table
\ref{tab:CurveLanes}. It can be found the SOTA methods such as SCNN
and SAD suffer substantial performance drop (20\%\textasciitilde 30\%
F1 score). Our method is lane sensitive and performs far better than
all the competing methods e.g. CurveLane-S can reach 81.12\% with
an F1 measure which is 16\% higher than the SCNN. We also show some
qualitative results on CurveLanes in Figure \ref{fig:Results-of-CurveLanes}.
Better performance of CurveLane-NAS can be found in the difficult
scenarios such as large-curves, night and wet roads. More qualitative
comparisons can be found in the Appendix.

The searched architectures also show strong transferability in dealing
with other lane detection tasks. We transfer the searched architecture
to the TuSimple dataset. With the optimal architecture searched in
CUlane, our method reached a comparable performance with SCNN but
much faster. The detailed tables can be found in the Appendix.

\textbf{Searched Architectures.} The detailed searched architectures
of CurveLane-NAS can be found in the supplementary materials. The
architecture is quite different from the hand-craft design such as
ResNet50. The backbone of the found architectures usually down-samples
twice (only 3 stages). The positions of doubling channels are usually
in the later stages to control the total FLOPS since larger channels
in the early stage which will result in a great computational burden.
Larger models use more heads. Most results of fusion module tend to
select output features from the first stage and the second stage with
the output one to enable more spatial information.

\textbf{Ablative Study.} We conduct ablation analysis of our proposed
model modifications in Table \ref{tab:Ablative-Study-onNMS}. The
study is based on different backbones (searched architectures and
the ResNet101) on the CULane dataset. It can be found that multi-level
heads are more useful for small models. Adaptive Points Blending modules
can boost the performance more the larger models. Figure \ref{fig:Performance-of-the-NMS-SWAP}
(Right) further shows some qualitative comparisons, it can be found
that our adaptive points blending can yield significantly better performance
than Plain-NMS for the curve lanes and remote part of the lanes.

\begin{table}[tb]

\caption{\label{tab:Ablative-Study-onNMS}Ablative study with the F1-measure
on the CULane dataset. CurveLane\_S to L denote our searched backbone
architectures. The performance of models combined with all the modules
are listed in the final column.}

\begin{centering}
{\scriptsize{}}\tabcolsep 0.02in{\scriptsize{}}%
\begin{tabular}{c|c|c|c|c}
\hline 
{\scriptsize{}Backbone} & {\scriptsize{}Backbone Only} & {\scriptsize{}+Feature Fusion} & {\scriptsize{}+Multi Level Heads} & {\scriptsize{}+Adaptive Points Blending}\tabularnewline
\hline 
{\scriptsize{}ResNet101} & {\scriptsize{}70.2} & {\scriptsize{}70.4$^{+0.2}$} & {\scriptsize{}71.9$^{+1.5}$} & {\scriptsize{}72.4$^{+0.5}$}\tabularnewline
{\scriptsize{}CurveLane\_S} & {\scriptsize{}69.5} & {\scriptsize{}70.1$^{+0.6}$} & {\scriptsize{}70.9$^{+0.8}$} & {\scriptsize{}71.5$^{+0.6}$}\tabularnewline
{\scriptsize{}CurveLane\_M} & {\scriptsize{}71.7} & {\scriptsize{}72.0$^{+0.3}$} & {\scriptsize{}72.2$^{+0.2}$} & {\scriptsize{}73.5$^{+1.3}$}\tabularnewline
{\scriptsize{}CurveLane\_L} & {\scriptsize{}72.6} & {\scriptsize{}73.1$^{+0.5}$} & {\scriptsize{}73.5$^{+0.4}$} & {\scriptsize{}74.8$^{+1.3}$}\tabularnewline
\hline 
\end{tabular}{\scriptsize\par}
\par\end{centering}

\end{table}

\section{Conclusion}

We propose CurveLane-NAS, a NAS pipeline unifying lane-sensitive architecture
search and adaptive point blending for curve lane detection. The new
framework can automatically fuse and capture both long-ranged coherent
and accurate curve information and enable a more efficient computational
allocation. The searched networks achieve state-of-the-art speed/FLOPS
trade-off comparing to existing methods. Furthermore, we release a
new largest lane detection dataset named CurveLanes for the community
to establish a more challenging benchmark with more curve lanes/lanes per image. 

\bibliographystyle{splncs04}
\bibliography{Know_network}

\begin{thebibliography}{10}
\providecommand{\url}[1]{\texttt{#1}}
\providecommand{\urlprefix}{URL }
\providecommand{\doi}[1]{https://doi.org/#1}

\bibitem{baker2016designing}
Baker, B., Gupta, O., Naik, N., Raskar, R.: Designing neural network
  architectures using reinforcement learning. In: ICLR (2017)

\bibitem{cai2018efficient}
Cai, H., Chen, T., Zhang, W., Yu, Y., Wang, J.: Efficient architecture search
  by network transformation. In: AAAI (2018)

\bibitem{cai2018proxylessnas}
Cai, H., Zhu, L., Han, S.: Proxylessnas: Direct neural architecture search on
  target task and hardware. In: ICLR2019 (2019)

\bibitem{chen2018searching}
Chen, L.C., Collins, M., Zhu, Y., Papandreou, G., Zoph, B., Schroff, F., Adam,
  H., Shlens, J.: Searching for efficient multi-scale architectures for dense
  image prediction. In: NeurIPS (2018)

\bibitem{chen2019detnas}
Chen, Y., Yang, T., Zhang, X., Meng, G., Pan, C., Sun, J.: Detnas: Neural
  architecture search on object detection. In: NeurIPS (2019)

\bibitem{chen2019pointlanenet}
Chen, Z., Liu, Q., Lian, C.: Pointlanenet: Efficient end-to-end cnns for
  accurate real-time lane detection. In: IV. pp. 2563--2568. IEEE (2019)

\bibitem{chiu2005lane}
Chiu, K.Y., Lin, S.F.: Lane detection using color-based segmentation. In: IV.
  pp. 706--711. IEEE (2005)

\bibitem{gonzalez2000lane}
Gonzalez, J.P., Ozguner, U.: Lane detection using histogram-based segmentation
  and decision trees. In: ITSC2000. 2000 IEEE Intelligent Transportation
  Systems. Proceedings (Cat. No. 00TH8493). pp. 346--351. IEEE (2000)

\bibitem{he2016deep}
He, K., Zhang, X., Ren, S., Sun, J.: Deep residual learning for image
  recognition. In: CVPR (2016)

\bibitem{DBLP:journals/corr/abs-1905-03704}
Hou, Y.: Agnostic lane detection. CoRR  (2019),
  \url{http://arxiv.org/abs/1905.03704}

\bibitem{hou2019learning}
Hou, Y., Ma, Z., Liu, C., Loy, C.C.: Learning lightweight lane detection cnns
  by self attention distillation. In: ICCV2019 (2019)

\bibitem{jiang2020sp}
Jiang, C., Xu, H., Zhang, W., Liang, X., Li, Z.: Sp-nas: Serial-to-parallel
  backbone search for object detection. In: CVPR. pp. 11863--11872 (2020)

\bibitem{lee2009effective}
Lee, J.W., Cho, J.S.: Effective lane detection and tracking method using
  statistical modeling of color and lane edge-orientation. In: 2009 Fourth
  International Conference on Computer Sciences and Convergence Information
  Technology. pp. 1586--1591. IEEE (2009)

\bibitem{li2019line}
Li, X., Li, J., Hu, X., Yang, J.: Line-cnn: End-to-end traffic line detection
  with line proposal unit. IEEE Transactions on Intelligent Transportation
  Systems  (2019)

\bibitem{liu2019auto}
Liu, C., Chen, L.C., Schroff, F., Adam, H., Hua, W., Yuille, A., Fei-Fei, L.:
  Auto-deeplab: Hierarchical neural architecture search for semantic image
  segmentation. In: CVPR (2019)

\bibitem{liu2018progressive}
Liu, C., Zoph, B., Neumann, M., Shlens, J., Hua, W., Li, L.J., Fei-Fei, L.,
  Yuille, A., Huang, J., Murphy, K.: Progressive neural architecture search.
  In: ECCV (2018)

\bibitem{liu2017hierarchical}
Liu, H., Simonyan, K., Vinyals, O., Fernando, C., Kavukcuoglu, K.: Hierarchical
  representations for efficient architecture search. In: ICLR (2018)

\bibitem{liu2018darts}
Liu, H., Simonyan, K., Yang, Y.: Darts: Differentiable architecture search. In:
  ICLR (2018)

\bibitem{DBLP:journals/corr/abs-1909-00798}
Mamidala, R.S., Uthkota, U., Shankar, M.B., Antony, A.J., Narasimhadhan, A.V.:
  Dynamic approach for lane detection using google street view and {CNN}. CoRR
  (2019), \url{http://arxiv.org/abs/1909.00798}

\bibitem{pan2018spatial}
Pan, X., Shi, J., Luo, P., Wang, X., Tang, X.: Spatial as deep: Spatial cnn for
  traffic scene understanding. In: AAAI (2018)

\bibitem{DBLP:journals/corr/abs-1907-01294}
Pizzati, F., Allodi, M., Barrera, A., Garc{\'{\i}}a, F.: Lane detection and
  classification using cascaded cnns. CoRR  (2019),
  \url{http://arxiv.org/abs/1907.01294}

\bibitem{real2019regularized}
Real, E., Aggarwal, A., Huang, Y., Le, Q.V.: Regularized evolution for image
  classifier architecture search. In: AAAI. vol.~33, pp. 4780--4789 (2019)

\bibitem{real2017large}
Real, E., Moore, S., Selle, A., Saxena, S., Suematsu, Y.L., Tan, J., Le, Q.V.,
  Kurakin, A.: Large-scale evolution of image classifiers. In: ICML (2017)

\bibitem{sandler2018mobilenetv2}
Sandler, M., Howard, A., Zhu, M., Zhmoginov, A., Chen, L.C.: Mobilenetv2:
  Inverted residuals and linear bottlenecks. In: CVPR. pp. 4510--4520 (2018)

\bibitem{szegedy2015going}
Szegedy, C., Liu, W., Jia, Y., Sermanet, P., Reed, S., Anguelov, D., Erhan, D.,
  Vanhoucke, V., Rabinovich, A.: Going deeper with convolutions. In: CVPR.
  pp.~1--9 (2015)

\bibitem{tan2018mnasnet}
Tan, M., Chen, B., Pang, R., Vasudevan, V., Le, Q.V.: Mnasnet: Platform-aware
  neural architecture search for mobile. In: CVPR (2019)

\bibitem{tan2019efficientnet}
Tan, M., Le, Q.V.: Efficientnet: Rethinking model scaling for convolutional
  neural networks. In: ICML (2019)

\bibitem{TuSimple}
TuSimple: Tusimple lane detection challenge. In: CVPR Workshops (2017)

\bibitem{xie2018snas}
Xie, S., Zheng, H., Liu, C., Lin, L.: Snas: stochastic neural architecture
  search. In: ICLR (2019)

\bibitem{xu2019auto}
Xu, H., Yao, L., Zhang, W., Liang, X., Li, Z.: Auto-fpn: Automatic network
  architecture adaptation for object detection beyond classification. In: ICCV
  (2019)

\bibitem{yao2020sm}
Yao, L., Xu, H., Zhang, W., Liang, X., Li, Z.: Sm-nas: Structural-to-modular
  neural architecture search for object detection. In: AAAI (2020)

\bibitem{yu2018bdd100k}
Yu, F., Xian, W., Chen, Y., Liu, F., Liao, M., Madhavan, V., Darrell, T.:
  Bdd100k: A diverse driving video database with scalable annotation tooling.
  In: CVPR (2020)

\bibitem{zhong2018practical}
Zhong, Z., Yan, J., Wu, W., Shao, J., Liu, C.L.: Practical block-wise neural
  network architecture generation. In: CVPR (2018)

\bibitem{zoph2016neural}
Zoph, B., Le, Q.V.: Neural architecture search with reinforcement learning. In:
  ICLR (2017)

\bibitem{zoph2018learning}
Zoph, B., Vasudevan, V., Shlens, J., Le, Q.V.: Learning transferable
  architectures for scalable image recognition. In: CVPR (2018)

\bibitem{zou2019robust}
Zou, Q., Jiang, H., Dai, Q., Yue, Y., Chen, L., Wang, Q.: Robust lane detection
  from continuous driving scenes using deep neural networks. arXiv preprint
  arXiv:1903.02193  (2019)

\end{thebibliography}
 
\end{document}